%% file: main.tex
\documentclass[10pt,twocolumn,letterpaper]{article}
\pdfoutput=1

\usepackage[pagenumbers]{cvpr} % To force page numbers, e.g. for an arXiv version

\input{preamble}
\definecolor{cvprblue}{rgb}{0.21,0.49,0.74}
\usepackage[pagebackref,breaklinks,colorlinks,allcolors=cvprblue]{hyperref}
\usepackage{lineno}
\usepackage{multirow}
\usepackage{amsmath}
\usepackage{bm}
\usepackage{xcolor}
\usepackage[table]{xcolor}
\usepackage{changepage}
\usepackage{pgfplots}
\usepackage{pgfplotstable}
\pgfplotsset{compat=1.18}
\usepgfplotslibrary{groupplots}
\usepgfplotslibrary{colorbrewer}
\usepgfplotslibrary{statistics}
\usepgfplotslibrary{external} 
\definecolor{darkgreen}{rgb}{0.0, 0.5, 0.0}
\definecolor{moderategrey}{rgb}{0.8, 0.8, 0.8} % RGB for light grey
\definecolor{lightgreen}{rgb}{0.6, 0.89, 0.6}

\def\paperID{4} % *** Enter the Paper ID here
\def\confName{CVPR}
\def\confYear{2026}

\title{Bringing a Personal Point of View: \\ Evaluating Dynamic 3D Gaussian Splatting for Egocentric Scene Reconstruction}

\author{
    Jan Warchocki\textsuperscript{1} \qquad
    Xi Wang\textsuperscript{2,3} \qquad
    Jonas Kulhanek\textsuperscript{4} \qquad
    Jan van Gemert\textsuperscript{1} \\
    \textsuperscript{1}Delft University of Technology \qquad
    \textsuperscript{2}ETH Z\"urich \qquad
    \textsuperscript{3}TUM/MCML \\
    \textsuperscript{4}Czech Technical University in Prague \\
    {\tt\small \{janwarchocki, jonas.kulhanek\}@gmail.com, xi.wang@inf.ethz.ch, j.c.vangemert@tudelft.nl}
}

\newcommand\notsotiny{\@setfontsize\notsotiny{5.5}{6.5}}
\newcommand{\Mod}[1]{\ (\mathrm{mod}\ #1)}
\def\modelcolormap{%
    EgoGaussian/purple,%
    Def3DGS/blue,%
    4DGS/orange,%
    RTGS/darkgreen}
\def\modelcolormapnoeg{%
    Def3DGS/blue,%
    4DGS/orange,%
    RTGS/darkgreen}
\pgfdeclarelayer{background layer}
\pgfdeclarelayer{foreground layer}
\pgfsetlayers{background layer,main,foreground layer}

\begin{document}
\maketitle
\input{sec/0_abstract}
\input{sec/1_introduction}
\input{sec/2_related_work}
% \input{sec/3_method}
\input{sec/4_results}
\input{sec/5_discussion}

\input{sec/6_conclusion}
{
    \small
    \bibliographystyle{ieeenat_fullname}
    \bibliography{main}
}

\input{sec/X_suppl}

\end{document}

%% file: sec/0_abstract.tex
\begin{abstract}
Egocentric video provides a unique view into human perception and interaction, with growing relevance for augmented reality, robotics, and assistive technologies. However, rapid camera motion and complex scene dynamics pose major challenges for 3D reconstruction from this perspective. While 3D Gaussian Splatting (3DGS) has become a state-of-the-art method for efficient, high-quality novel view synthesis, variants, that focus on reconstructing dynamic scenes from monocular video are rarely evaluated on egocentric video. It remains unclear whether existing models generalize to this setting or if egocentric-specific solutions are needed. In this work, we evaluate dynamic monocular 3DGS models on egocentric and exocentric video using paired ego-exo recordings from the EgoExo4D dataset. We find that reconstruction quality is consistently lower in egocentric views. Analysis reveals that the difference in reconstruction quality, measured in peak signal-to-noise ratio, stems from the reconstruction of static, not dynamic, content. Our findings underscore current limitations and motivate the development of egocentric-specific approaches, while also highlighting the value of separately evaluating static and dynamic regions of a video.
\end{abstract}

%TODOk
%While our exocentric evaluation reuses static camera viewpoints across time, potentially simplifying the task, our results still highlight the additional generalization demands posed by egocentric video.

%% file: sec/1_introduction.tex
\section{Introduction}
\label{sec:intro}

%Egocentric, or first-person, video captures the visual input received by an agent acting in the world, such as a human wearing a head-mounted camera. This agent-centric perspective is especially relevant for robotics, where understanding and interacting with the environment from a first-person view is essential. Egocentric data provides a natural foundation for training models that perceive and act based on what the agent sees, making it valuable for applications in robotics \cite{robotic_hand_ego}, assistive technologies \cite{assistive_corobot}, and augmented reality \cite{ar_egocentric}. Unlike third-person recordings, egocentric video reflects the sensory input available to an agent during decision-making and action. Leveraging such data can lead to more intuitive and effective behaviors in real-world scenarios \cite{assistive_corobot}.

Egocentric, or first-person, video captures the visual input received by an agent acting in the world, such as a human wearing a head-mounted camera. This type of data provides a natural window into how humans perceive and interact with their surroundings, making it especially valuable for applications in augmented reality \cite{ar_egocentric}, assistive technology \cite{assistive_corobot}, and robotics \cite{robotic_hand_ego}. Unlike third-person recordings, egocentric video closely reflects the visual input an agent receives while acting in the world. Using egocentric data, we can improve an agent's ability to understand the world from a first-person perspective, leading to more intuitive and efficient interactions in real-world scenarios \cite{assistive_corobot}.

Egocentric data presents unique challenges. The camera is subject to rapid, often unpredictable motion driven by head or body movement \cite{epic_fields, ego_slam, egocentric_visual_intention}. At the same time, the scenes themselves are highly dynamic, frequently involving complex hand-object interactions \cite{epic_fields, EgoGaussian, ego_lifter}. To address these challenges, large-scale egocentric datasets such as EPIC-Kitchens \cite{epic_kitchens, epic_kitchens_55}, HOI4D \cite{hoi4d}, and Ego4D~\cite{ego4d} have driven advances in tasks like action recognition \cite{lsta}, action prediction \cite{egocentric_action_prediction}, and hand-object segmentation \cite{egohos}. Yet, one important direction remains largely underexplored: 3D reconstruction and novel view synthesis in egocentric settings. Tackling this gap is key to enabling more comprehensive spatial understanding and unlocking immersive applications such as augmented and virtual reality \cite{ego_lifter}.

3D Gaussian Splatting (3DGS) \cite{3dgs} has recently emerged as a state-of-the-art approach for high-quality and efficient 3D reconstruction and novel view synthesis with use cases in fields such as robotics \cite{mani_gaussian}. Extensions have been proposed to the 3DGS framework to allow the handling of dynamic scenes viewed from a single camera (monocular) \cite{def3dgs, 4dgs, rtgs, EgoGaussian, 4d_rotor, wasserstein_gaussians, gaufre}. These methods, however, are commonly only evaluated on scenes filmed from a third-person, exocentric perspective, rather than from the egocentric perspective \cite{monocular_evaluation, d_nerf, dycheck, nerfies, nerf_ds}. Hence, it remains unclear how well these models perform when applied to egocentric video, and whether models specialized on the egocentric perspective, are needed.

To our knowledge, EgoGaussian \cite{EgoGaussian} and DeGauss \cite{degauss} are the only dynamic monocular 3DGS models specifically designed for egocentric vision. While the authors of EgoGaussian \cite{EgoGaussian} report improved rendering quality over monocular baselines \cite{def3dgs, 4dgs} on egocentric data, we were unable to reproduce these gains in our experiments. On the other hand, no quantitative evaluation on egocentric performance was performed for DeGauss \cite{degauss}. This leaves open the question of whether egocentric-specific models offer a clear advantage.

In this work, we aim to address this research gap and answer the question of how well existing monocular dynamic 3D Gaussian Splatting models perform when applied to egocentric data. To this end, we compare the existing models on paired ego and exo perspective recordings of the same scene from the EgoExo4D dataset \cite{ego_exo}. Because dynamic regions may be crucial in practical applications \cite{endo_4dgs}, we compare the performance of the models on the static and dynamic regions separately. Since rapid camera motion is often cited as a challenge of egocentric vision \cite{epic_fields, ego_slam}, we investigate if it correlates with reconstruction quality. Overall, our main contributions can be summarized as follows: 

\begin{enumerate}
    \item We compare the performance of four existing dynamic 3DGS models on paired ego-exo scenes from the EgoExo4D dataset.
    \item We propose an evaluation protocol that compares the performance of existing models on static and dynamic regions of the scene.
    \item We propose methods to study the correlation between camera motion and 3D reconstruction quality in egocentric settings.
    \item Our results suggest the need for egocentric-specific approaches, while also showing that future evaluations of methods could benefit from evaluating the static and dynamic regions of the scenes separately.
\end{enumerate}

Our code has been made available online at \href{https://github.com/Jaswar/evaluating-3dgs-egocentric}{https://github.com/Jaswar/evaluating-3dgs-egocentric}. 

%\textbf{1)} We compare the performance of four existing dynamic 3DGS models on paired ego-exo scenes from the EgoExo4D dataset. \textbf{2)} We propose an evaluation protocol that compares the performance of existing models on static and dynamic regions of the scene. \textbf{3)} We study the correlation between camera motion and reconstruction quality in egocentric settings. \textbf{4)} Our results suggest the need for egocentric-specific approaches, while also showing that future evaluations of methods could benefit from evaluating the static and dynamic regions of the scenes separately. Our code and data are available at \href{https://github.com/Jaswar/evaluation-thesis}{https://github.com/Jaswar/evaluation-thesis}. 

%% file: sec/2_related_work.tex
\section{Related work}
\label{sec:related}

\textbf{Egocentric vision.} Egocentric, or first-person vision, focuses on capturing visual data from the viewpoint of a wearable or head-mounted camera. This special type of vision has recently garnered attention due to its importance for applications such as augmented reality \cite{ar_egocentric} and robotics~\cite{assistive_corobot, robotic_hand_ego}. Although egocentric video is considered difficult due to issues such as the complexity of human actions \cite{ego_lifter, vscos} and varied camera motion \cite{epic_fields, ego_slam}, egocentric datasets such as Epic-Kitchens \cite{epic_kitchens, epic_kitchens_55} have helped advancements in fields such as video understanding and human-object interaction \cite{visor, egocentric_action_prediction}. In this work, we compare egocentric recordings to other types of recordings to verify whether the challenges posed by egocentric vision impact the performance of monocular 3D Gaussian Splatting methods. 

%First person, egocentric, vision has been of interest for fields such as augmented reality (AR) \cite{hoi4d} and robotics \cite{ego_exo, hoi4d}. Datasets, such as Epic-Kitchens \cite{epic_kitchens, epic_kitchens_55}, HOI4D \cite{hoi4d}, and Ego4D \cite{ego4d} caused advances in tasks such as video understanding and hand-object segmentation. However, the field of 3D Gaussian splatting from egocentric videos remains largely unexplored. To our knowledge, EgoGaussian \cite{EgoGaussian} is currently the only model focused on reconstructing 3D scenes from dynamic, egocentric videos. In this work, we compare EgoGaussian with other existing monocular models in egocentric settings. 

\textbf{3D Gaussian Splatting for dynamic scenes.} 3D Gaussian Splatting (3DGS) \cite{3dgs} has recently emerged as a promising method for novel view synthesis of static scenes, outperforming existing NeRF-based approaches \cite{nerf, mip_nerf} in both rendering quality and speed \cite{3dgs}. 3DGS assumes the scene to be static, leading to artefacts, such as floaters, in the reconstruction of dynamic scenes \cite{EgoGaussian, ego_lifter}. As such, special methods have been proposed to handle scene motion \cite{dyn3dgs, wasserstein_gaussians, relay_gs}. Of these, monocular methods \cite{def3dgs, 4dgs, rtgs, 4d_rotor, gaufre, gaussian_flow, spline_gs}, which require only a single camera, are especially interesting \cite{monocular_evaluation}. In this work, we focus on evaluating monocular models in egocentric dynamic scenes. 

% 3D Gaussian splatting (3DGS) \cite{3dgs} assumes a static scene, leading to artefacts, such as floaters, present in the reconstruction of dynamic scenes \cite{EgoGaussian, ego_lifter}. As such, special methods have been proposed to handle scene motion \cite{dyn3dgs, wasserstein_gaussians, relay_gs}. Of these, monocular methods \cite{def3dgs, 4dgs, rtgs, 4d_rotor, gaufre, gaussian_flow, spline_gs}, which require only a single viewpoint, are especially interesting \cite{monocular_evaluation}. Although these methods have shown strong performance on exocentric, third-person data \cite{monocular_evaluation}, it is currently unknown how well they perform in egocentric, first-person settings. In this work, we aim to benchmark existing monocular 3DGS models on dynamic, egocentric views and compare their performance on the same scenes viewed from an exocentric angle. 

\textbf{Evaluation of existing monocular models.} Various datasets are used for the evaluation of models for 3D Gaussian Splatting from monocular videos with dynamics~\cite{monocular_evaluation, d_nerf, nerfies, hyper_nerf, nerf_ds, dycheck}. D-Nerf \cite{d_nerf} contains synthetic scenes captured with a rapidly moving camera without motion blur. Nerfies \cite{nerfies}, HyperNerf \cite{hyper_nerf}, and DyCheck \cite{dycheck} all contain real-world recordings of kitchen activities, animals, and other moving objects. However, even the scenes involving human actors are not recorded from an egocentric perspective, but rather from an exocentric, third-person perspective. As such, it is currently unknown how well existing monocular 3DGS models perform when the recording is captured from an egocentric point of view and whether these models perform better or worse than with other types of recordings. In this work, we provide such an evaluation. 

\textbf{EgoGaussian and DeGauss.} To the best of our knowledge, EgoGaussian \cite{EgoGaussian} was the first publicly-available model for monocular 3DGS reconstruction of dynamic scenes from an egocentric perspective. The model requires each clip to be manually split into passive (no interaction) and active (object manipulation) segments\footnote{Referred to as \textit{static} and \textit{dynamic} in the original paper; we adopt different terms to avoid confusion. In EgoGaussian \cite{EgoGaussian}, a \textit{dynamic} segment is a frame sequence with at least one moving object. Here, \textit{dynamics} refers to binary masks of moving objects. Same change in terms applies to statics.}. Using provided object masks, passive segments are then used to initialize the background and object shape, while active segments are used to refine both and estimate the object's pose. The method assumes fully rigid object motion and does not model the actor. We compare its reconstruction quality to monocular models not tailored to egocentric settings. 

Concurrent with our work, DeGauss \cite{degauss} has been introduced as an alternative dynamic 3DGS model tailored to the egocentric perspective. Unlike EgoGaussian \cite{EgoGaussian} that relies on provided object masks, DeGauss learns to segment dynamic regions directly, without requiring a split between passive and active segments or restricting motion to be rigid. The method has so far been evaluated qualitatively on egocentric data, and quantitative performance relative to baselines remains to be established. A public implementation was released after the completion of our experiments, so we leave the quantitative comparison to future work.

\input{figures/dataset_selected}

\textbf{Static and dynamic modeling.} Modeling dynamic objects is crucial for practical applications \cite{endo_4dgs}. Although the majority of dynamic and monocular 3DGS methods model static and dynamic regions of the video together~\cite{def3dgs, 4dgs, rtgs, wasserstein_gaussians, 4d_rotor}, methods exist where the static and dynamic are being modeled separately \cite{EgoGaussian, gaufre}. In \cite{monocular_evaluation} the authors show, however, that existing methods reconstruct the static and dynamic regions similarly. Since the evaluation in \cite{monocular_evaluation} focuses on non-egocentric data, we perform a similar analysis in egocentric settings. In this way, we aim to show whether future models could benefit from modeling the static and dynamic regions separately. 

%\textbf{Simultaneous Localization and Mapping (SLAM).} SLAM methods estimate an agent’s trajectory (localization) while building a map of the environment (mapping)~\cite{slam_survey}. Unlike 3DGS, which typically assumes known camera poses obtained offline, SLAM estimates poses online and often runs in real-time. Extensions have been proposed to handle dynamic environments \cite{slam_telepresence, dynamic_3dgs_slam, wildgs_slam} while Ego-SLAM \cite{ego_slam} showed the need for egocentric-specific approaches. While this work focuses on 3DGS, recent research integrates 3DGS into SLAM pipelines \cite{dynamic_3dgs_slam, wildgs_slam}, suggesting our findings could also inform SLAM. Separately, fast view synthesis methods like FaDIV-Syn \cite{fadiv_syn} offer an alternative by generating novel views without full 3D reconstruction, differing from the approaches studied here.

% \footnote{Referred to as \textit{static} and \textit{dynamic} in the original paper; we adopt different terms to avoid confusion.}

%% file: figures/dataset_selected.tex
\begin{figure*}[t!]
    \centering
    \includegraphics[width=\linewidth]{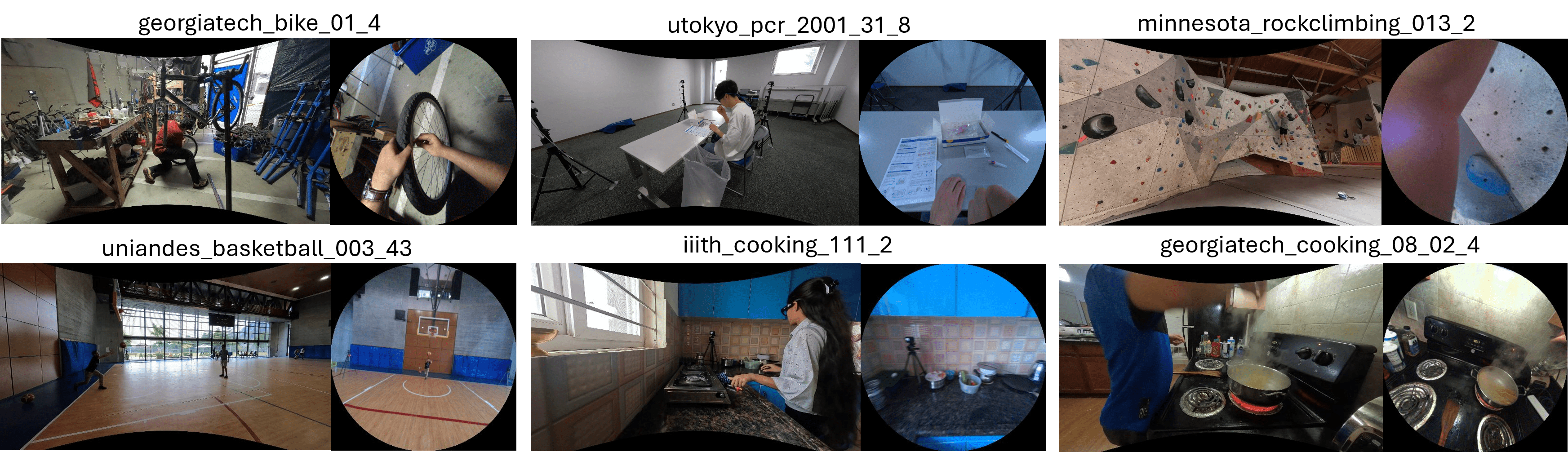}
    \caption{First frame of selected scenes from the exo (left, first exo camera) and ego (right) views. First 4 scenes are random; last 2 are selected EgoGaussian-style scenes. Frames shown after undistortion. As we can see, the scenes are varied. Best viewed zoomed in.}
    \label{fig:dataset_selected}
\end{figure*}

%% file: sec/4_results.tex
\section{Experiments}
\label{sec:results}

\subsection{Ego vs.\ exo}

In this experimental section, we aim to answer whether reconstruction from egocentric data is indeed different from other types of data. To this end, we compare the performance of existing models on egocentric data against the performance on exocentric, third-person data. We choose to compare against exocentric recordings as they form a natural opposite of egocentric recordings.

\textbf{Dataset.} EgoExo4D \cite{ego_exo}, contains 1,286 hours of paired egocentric and exocentric recordings of skilled human activities. We choose it over other ego-exo datasets~\cite{charades_ego, assembly101, h2o, home_action_genome, cmu_mmac, ego_exo_learn, castle} due to its scene diversity and availability of ground truth camera intrinsics, extrinsics, and a semi-dense 3D point cloud for initializing Gaussians. These ground truth parameters are crucial for isolating baseline performance from errors introduced by structure-from-motion methods such as COLMAP \cite{colmap, EgoGaussian}.

We first sample 20 clips at random, each exactly 300 frames (10 seconds) long, which approximately corresponds to the length of existing dynamic clips for the task of monocular 3DGS of dynamic scenes \cite{EgoGaussian}. To evaluate EgoGaussian \cite{EgoGaussian}, additional clips must be selected manually to fulfill the requirements of EgoGaussian: rigid motion and a split into passive and active segments \cite{EgoGaussian}. To this end, we selected 5 such clips. In total, the dataset therefore contains 25 scenes and the first frames from both exo and ego views of selected scenes are presented in \Cref{fig:dataset_selected}. As we can see, the clips contain diverse scenarios. The exo cameras tend to capture more depth information about the scene and the exo cameras are static. These differences may cause the scene to be reconstructed with a different quality from the ego and exo perspectives. More example scenes are visualized in Supplementary C. 

\textbf{Models.} Apart from EgoGaussian, we select three other baseline methods for monocular 3D Gaussian Splatting of dynamics scenes. The baselines are: Deformable-3DGS~\cite{def3dgs}, 4DGS \cite{4dgs}, and RTGS \cite{rtgs}. We select Deformable-3DGS and 4DGS due to their strong performance on dynamic scenes as shown by \cite{monocular_evaluation}. Since both Deformable-3DGS and 4DGS define the motion with a global field \cite{def3dgs, 4dgs, monocular_evaluation}, we include RTGS for model variety, as it does not rely on an explicit motion field~\cite{rtgs, monocular_evaluation}.

\textbf{Data preprocessing.} The ego and exo cameras in Ego-Exo4D are fisheye and incompatible with the standard 3D Gaussian Splatting framework \cite{ego_exo, 3dgs, fisheye_gs}. We address this by undistorting frames using known intrinsics, following the official guidelines \cite{ego_exo_undistortion}. This process introduces artifacts, producing black regions at the top and bottom of exo images \cite{ego_exo_undistortion}, as shown in \Cref{fig:dataset_selected}. To filter invalid pixels, we undistort a binary mask alongside the images and use it within the 3D Gaussian Splatting pipeline. 

We manually provide the split into passive and active segments for EgoGaussian. Furthermore, we obtain the object and actor masks by annotating frames using Segment Anything 2 \cite{sam2}.

\input{figures/renderings}

\textbf{Evaluation protocol.} We compare model performance on time-synchronized ego and exo recordings of the same scene. For ego views, the training set uses even-indexed frames, validation uses frames where index $i \equiv 1 \Mod{4}$, and test where $i \equiv 3 \Mod{4}$. This mirrors EgoGaussian~\cite{EgoGaussian}, but with a validation split added for per-scene hyperparameter tuning.

Since exo cameras are static, a single-view recording lacks sufficient multi-view information for training 3DGS models \cite{3dgs_from_single_image}. Instead, we generate the sequence by randomly selecting a viewpoint at each index $i$. This ensures monocular input while preserving 3D cues, and is similar to existing setups \cite{d_nerf, hyper_nerf, nerfies}. We apply the same train, validation, and test split as in the egocentric case. 

\textbf{Evaluation metrics.} Following \cite{monocular_evaluation}, we report masked peak signal-to-noise ratio (mPSNR), masked structural similarity index measure (mSSIM), and masked learned perceptual image patch similarity (mLPIPS). For the eight random scenes, the mask corresponds to the undistorted binary mask defined earlier. For the EgoGaussian scenes, the mask further excludes the actor, as it is not within the reconstruction scope of EgoGaussian \cite{EgoGaussian}. All metrics are computed on the test set. Each model is re-trained 3 times per scene using the same hyperparameters.

\textbf{Hyperparameters.} We perform a random search to select per-scene hyperparameters for Deformable-3DGS, 4DGS, and RTGS. For 4DGS and RTGS, the search space includes parameter values from existing configurations. As Deformable-3DGS lacks such configurations, we instead search over the width and depth of the deformation network, as well as the total number of iterations. The exact search parameters for all models are provided in Supplementary A.

The search runs for a fixed duration of 4 hours per scene and per model. Each configuration is evaluated on the validation set, and the one with the highest PSNR is selected. All searches are performed on a single NVIDIA A40 GPU. No hyperparameter search was conducted for EgoGaussian as its training time exceeded 4 hours on each scene.

\input{tables/full_results}

\textbf{Results.} The results of this experiment are presented in \Cref{tab:full_results}, where the metrics are averaged over the 
20 random and 5 EgoGaussian scenes separately. As we can observe, the models almost always perform better on the exocentric recordings with very low variance in results. The only exception is the mPSNR score measured for the RTGS model on the EgoGaussian scenes. However, other metrics in this setting are still better for exocentric recordings. This difference in reconstruction quality between ego and exo suggests that the reconstruction from egocentric perspective is more difficult for existing models on average.

The gap in mPSNR performance between ego and exo appears larger in the random scenes than the EgoGaussian scenes. However, EgoGaussian clips consist of passive segments where no object interaction happens. Additionally, the hands are excluded from egocentric metric calculation and the object movement is fully rigid. These differences may cause the scene dynamics to be easier to reconstruct from the egocentric perspective, which could explain the mPSNR difference between ego and exo being smaller than in random scenes. This observation suggests that EgoGaussian clips may not be fully representative of models' performance on egocentric video. 

Comparing EgoGaussian to other models, we observe that it obtains worse reconstruction quality across all metrics. This is a surprising result, as in the original paper, EgoGaussian was shown to outperform both Deformable-3DGS and 4DGS on egocentric data \cite{EgoGaussian}. 

Example renderings are presented in \Cref{fig:renderings}. As we can see from the first two, random sequences, the overall reconstruction from the ego perspective is visibly worse than the exo view. In the basketball sequence, the difference is most visible due to the poor reconstruction of the basketball and the hands of the actor. Although the qualitative results reinforce the conclusion that the reconstruction from the egocentric perspective is more difficult, it is yet unclear whether this difference comes from the reconstruction of dynamic or static objects. 

% Finally, we note that the standard deviations 

\subsection{Dynamic vs.\ static}

\input{tables/dynamic_results}

Accurate dynamic modeling is essential for practical applications involving dynamic 3DGS \cite{endo_4dgs}. Additionally, dynamic objects pose different challenges than static objects and some methods model them separately \cite{EgoGaussian, gaufre}. In this section, we thus aim to answer whether current methods model static and dynamic regions with the same accuracy.

\textbf{Dynamic masks.} We use Segment Anything 2 \cite{sam2} to manually annotate each dynamic object in selected clips. The mask for a given object at a given frame $i$ is only considered dynamic if the object visibly moved between frames $i$ and $i - 1$. Example resulting dynamic masks have been overlaid in green in \Cref{fig:renderings}.

\textbf{Evaluation metrics.} Similarly to the previous section, we use mPSNR, mSSIM, and mLPIPS to evaluate the models. The dynamic mask corresponds to the combined masks of all dynamic objects at the given frame. The static mask contains only the background, \ie all objects that are not currently moving. The static and dynamic masks are combined with the undistortion masks obtained previously to ensure only valid pixels are evaluated. 

\textbf{Results.} The results for the dynamic and static masks are presented in \Cref{tab:dynamic_results}. Firstly, as we can see, the model variance remains low on both dynamic and static parts of the scene. Secondly, we observe that it is unclear whether the reconstruction of dynamics is easier from the ego or exocentric views. In random scenes, the dynamic mPSNR for Deformable-3DGS and 4DGS is higher for the egocentric view. The egocentric mPSNR for RTGS is also much closer to the exo perspective than in \Cref{tab:full_results}. At the same time, the static reconstruction is again of higher quality in terms of mPSNR in exocentric views. Both mSSIM and mLPIPS are better in the exocentric view on static and dynamic regions. Hence, these results suggest that the gap in reconstruction quality between ego and exo in terms of mPSNR comes from the reconstruction of static regions, not dynamic. 

In the EgoGaussian scenes, we observe that the dynamic reconstruction is almost always better in the egocentric case. This reinforces the previous hypothesis that the dynamics in EgoGaussian videos are easier to reconstruct in the egocentric view. Additionally, this further suggests that the EgoGaussian clips may not form a representative sample of real videos.

%While the model variance remains low, as we can see from \Cref{tab:dynamic_results}, the reconstruction of dynamic objects is no longer clearly better in the exocentric case. In random scenes, the mPSNR is higher for both Def3DGS and 4DGS in the ego case. In EgoGaussian scenes, egocentric reconstruction is easier on average with most models performing better on this type of data. On the other hand, the reconstruction of static objects (\Cref{tab:static_results}) is again easier for all models in the exocentric case. 

Furthermore, the difference in performance between static and dynamic reconstruction is clearly visible both from \Cref{tab:dynamic_results} as well as \Cref{fig:renderings}. Across both egocentric and exocentric recordings, the models perform better when reconstructing the static regions of the scene. This highlights a key limitation of current methods in handling motion and suggests that future work should focus on explicitly improving dynamic scene understanding. 

\input{tables/egogaussian_ek_no_sub_results}
As with the previous results, EgoGaussian again performs worse than other baselines. The performance is worse in both the static and dynamic regions. This is again unexpected considering the original results \cite{EgoGaussian}.

\subsection{EgoGaussian vs.\ others}
\label{sec:egogaussian_vs_others} 

As shown in Tables \ref{tab:full_results} and \ref{tab:dynamic_results}, our results for EgoGaussian differ from the original paper, where it outperformed 4DGS and Deformable-3DGS both quantitatively and qualitatively~\cite{EgoGaussian}. To validate our pipeline, we re-evaluate EgoGaussian and the baselines on the original EgoGaussian data from Epic-Kitchens \cite{epic_kitchens, epic_kitchens_55} and HOI4D \cite{hoi4d}.

\textbf{Evaluation protocol.} We maintain the evaluation protocol from EgoGaussian and hence only split the data into a train and a test set. It is unknown which hyperparameters were used for the baselines in the EgoGaussian paper \cite{EgoGaussian}, hence we use default configurations. Both for the baselines and EgoGaussian we do not measure the reconstruction quality of body parts. Therefore, we measure masked PSNR, SSIM, and LPIPS. 

Furthermore, it should be noted that in the official EgoGaussian evaluation, the masked out areas are zeroed, rather than ignored \cite{EgoGaussian}, which will lead to biased metric values. Additionally, EgoGaussian does not normalize input images to the $[-1, 1]$ range, which is necessary for LPIPS \cite{lpips}. We report EgoGaussian results with (ours$^\dagger$) and without (ours$^*$) correction in metrics. 

\textbf{Results.} The results are presented in \Cref{tab:egogaussian_ek_no_sub_results.tex}. As we can observe, EgoGaussian without metric changes (ours$^*$) closely matches the original paper. However, the baselines perform much better than originally reported. Indeed, when comparing with updated metric calculations (ours$^\dagger$), the baselines tend to outperform EgoGaussian itself. These results therefore reinforce the findings from the previous sections and show that the sudden difference in performance does not come from the data, preprocessing, or the lack of hyperparameter tuning for EgoGaussian.

%To better understand the issue, we contacted the authors of EgoGaussian, who kindly agreed to privately share the code used for the evaluation of Deformable-3DGS. Upon reviewing the code, we noticed what appears to be a bug related to the handling of timestamps when loading the model. We reached out to the authors to clarify whether this was indeed a bug, but did not receive a response.

% Additionally, it should be noted that we observed a common problem with both the Epic-Kitchens and HOI4D data where the frames may be duplicated due to poor lighting \cite{epic_kitchens_55}. This leads the testing set to be very similar to the training set. In turn, this might lead overfitting models to be preferred in the evaluation. We discuss this problem in more detail in ... . \todo{TODO: add this as an appendix/supplementary material.}

\subsection{Effects of camera motion}\label{sec:camera_motion_results}

Rapid and often unpredictable camera motion, caused by head or body movement, is an inherent property of egocentric data \cite{epic_fields}. Thus, understanding the impact or correlation of camera motion on reconstruction quality is essential. In this section, we aim to answer whether egocentric camera motion correlates with the reconstruction quality. 

\textbf{Definition of camera motion.} We measure camera motion in two aspects: camera velocity, defined as the speed of the camera between frames, and camera baseline, defined as the distance traveled by the camera. An increase in camera velocity causes the egocentric test camera poses to be further away from training poses, which may influence reconstruction quality. An increase in the camera baseline might provide more multi-view information, which could increase the quality \cite{monocular_evaluation, dycheck}. We note that although metrics have been proposed to measure the amount of multi-view information in a monocular setting \cite{dycheck}, these metrics are either difficult to compute in practice \cite{dycheck, monocular_evaluation}, or were already shown not to correspond well to actual reconstruction quality \cite{monocular_evaluation}. 

\subsubsection{Camera velocity}

%\textbf{Evaluation protocol.} Let $\mathbf{v}_t \in \mathbb{R}^3$ and $\bm{\omega}_t \in \mathbb{R}^3$ be the camera linear and angular velocities between time steps $t$ and $t-1$ stemming from camera translation and rotation. The maximal components of both velocities at a given time step $t$ are then computed as $\hat{v}_t = \max_{1 \leq i \leq 3} |\mathbf{v}_{ti}|$ and $\hat{\omega}_t = \max_{1 \leq i \leq 3} |\bm{\omega}_{ti}|$. Since the range of velocities varies in each scene, we normalize them to the $[0,1]$ range per scene. The velocities are plotted on a logarithmic scale for readability. Hence, the final linear and angular velocities $\bar{v}_t$ and $\bar{\omega}_t$ used for analysis are computed according to the equations:

% \begin{align}
%     \bar{v}_t &= \ln \left( \frac{\hat{v}_t - \min_t \left(\hat{v}_t\right)}{\max_t \left(\hat{v}_t\right) - \min_t \left(\hat{v}_t\right)} \right) \\
%     \bar{\omega}_t &= \ln \left( \frac{\hat{\omega}_t - \min_t \left(\hat{\omega}_t\right)}{\max_t \left(\hat{\omega}_t\right) - \min_t \left(\hat{\omega}_t\right)} \right)
% \end{align}

\textbf{Evaluation protocol.} Let $\mathbf{v}_t \in \mathbb{R}^3$ be the camera linear velocity between time steps $t$ and $t-1$ stemming from camera translation. The maximal components of the velocity at a given time step $t$ are then computed as $\hat{v}_t = \max_{1 \leq i \leq 3} |\mathbf{v}_{ti}|$. Since the range of velocities varies in each scene, we normalize them to the $[0,1]$ range per scene. The velocities are plotted on a logarithmic scale for readability. Hence, the final linear velocities $\bar{v}_t$ used for analysis are computed according to the equation:

\begin{align}
    \bar{v}_t &= \ln \left( \frac{\hat{v}_t - \min_t \left(\hat{v}_t\right)}{\max_t \left(\hat{v}_t\right) - \min_t \left(\hat{v}_t\right)} \right)
\end{align}

We then plot the linear velocities at time step $t$ against mLPIPS achieved by the model on the static part of the scene at the same time step. We choose to evaluate against LPIPS as it was shown to correspond with human perception better than PSNR and SSIM \cite{lpips}. Since mLPIPS also varies per scene, we normalize it to the $[0, 1]$ range, similarly to the velocity. Only test frames are evaluated.

\input{figures/linear_velocity}

\textbf{Results.} The resulting scatter plot for linear velocity is presented in \Cref{fig:linear_velocity}. Additionally, we plot trend lines by averaging the mLPIPS over buckets of size 0.05. Although initially the trend lines are noisy due to only few samples being available, after around $\bar{v}_t > 0.5$, as camera velocity increases, the average masked mLPIPS consistently increases, indicating worse reconstruction. This suggests that, contrary to some prior expectations \cite{dycheck}, increased camera movement in egocentric video does not always yield better reconstructions, and may in fact hinder performance. Further analysis in Supplementary D shows that the same correlation holds for increased camera rotation and when measuring mPSNR and mSSIM instead of mLPIPS.

Due to the high variance observed in the scatter plot, one might question whether the observed positive correlation is statistically significant. To this end, we measure the Pearson and Spearman coefficients. Both indicate a positive correlation of around 0.5 at a p-value, with the null hypothesis that the correlation is 0, far below 0.05. Supplementary D contains detailed results of the significance tests.

\subsubsection{Camera baseline}
\label{sec:results_camera_baseline}

%\textbf{Evaluation protocol.} Liang \etal \cite{monocular_evaluation} build a synthetic monocular dataset where the camera moves along an arch. The camera baseline is then defined as the distance between the start and end points of the camera. Since the trajectory in egocentric videos is more complex, we instead define the camera \textit{linear} baseline as the maximal distance between any two points on the camera trajectory. We compute the camera linear baseline per scene and report the mean mLPIPS over the test frames of the given sequence. EgoGaussian is not evaluated due to it only being tested on 5 scenes. 

\textbf{Evaluation protocol.} Liang \etal \cite{monocular_evaluation} built a synthetic monocular dataset with the camera moving along an arc, defining the baseline as the distance between the start and end points. Since the egocentric trajectories are more complex, we instead define the \textit{linear} baseline as the maximum distance between any two points on the trajectory. We compute this metric on a logarithmic scale per scene and report the mean mLPIPS over the sequence’s test frames. EgoGaussian is excluded, as it was only evaluated on 5 scenes.

\input{figures/linear_baseline}

\textbf{Results.} The results are shown in \Cref{fig:linear_baseline} where additional best-fit linear regression models are shown. As we can observe, there is no clear correlation on reconstruction quality. Although the best-fit linear regression lines have a positive slope, the correlations for Def3DGS and RTGS are not statistically significant when performing Pearson and Spearman tests. Importantly, however, no negative correlation is observed, which would signify that the reconstruction performance increases with the camera baseline. Together with the velocity analysis, these findings thus show that increased camera motion does not lead to an improvement in reconstruction and may, in fact, hinder it. This challenges assumptions from prior work \cite{dycheck, monocular_evaluation}.

Similarly to camera velocity, we can define a baseline based on angular motion, which we refer to as the \textit{angular} baseline. Supplementary D presents this additional analysis, showing that the same lack of negative correlation holds, and includes detailed significance test results.

%% file: figures/renderings.tex
\begin{figure*}[t!]
    \centering
    \includegraphics[width=\linewidth]{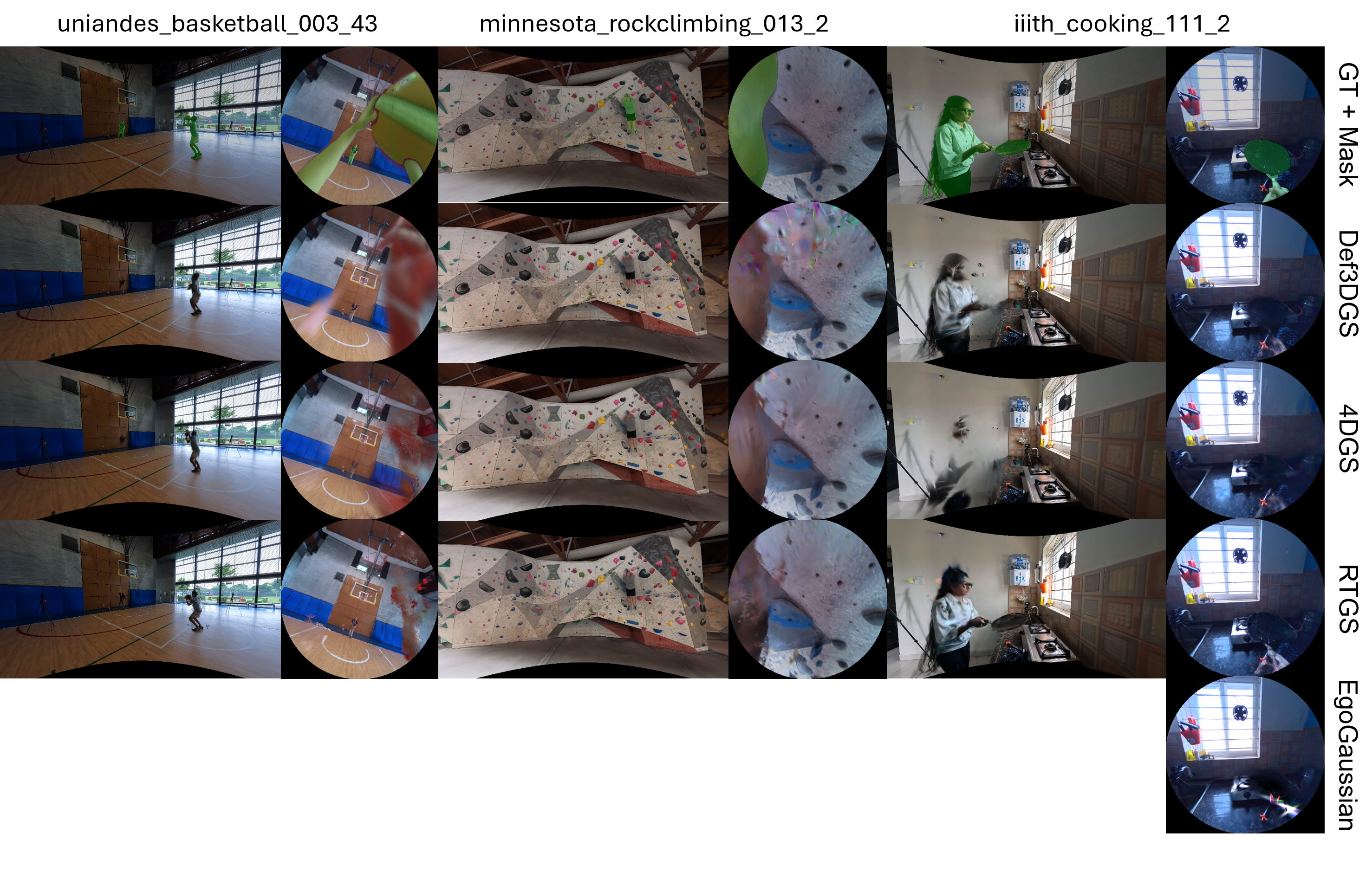}
    \caption{Ground truths and model renderings on 2 random and 1 EgoGaussian scene. EgoGaussian could only be evaluated on the ego view of one of these scenes due to the constraints imposed by the model. Top row shows ground truth with dynamic masks overlaid in green. As we can see, models consistently reconstruct dynamic regions with lower visual fidelity. Best viewed zoomed in.}
    \label{fig:renderings}
\end{figure*}

%% file: tables/full_results.tex
\begin{table}[t!]
    \tiny
    \setlength{\tabcolsep}{2pt}
    \centering
    \begin{tabular}{cccccccc}
        \cmidrule[0.5pt]{1-8}
         &  & \multicolumn{3}{c}{\textbf{Random scenes}} & \multicolumn{3}{c}{\textbf{EgoGaussian scenes}} \\
        \cmidrule(lr){3-5} \cmidrule(lr){6-8} 
        \textbf{Model} & \textbf{View} & mPSNR $\uparrow$ & mSSIM $\uparrow$ & mLPIPS $\downarrow$ & mPSNR $\uparrow$ & mSSIM $\uparrow$ & mLPIPS $\downarrow$ \\
        \cmidrule{1-8}
        EgoGaussian & Ego & - & - & - & 26.97 $\pm$ 0.04 & 0.77 $\pm$ 0.001 & 0.30 $\pm$ 0.001  \\
        \cmidrule{1-8}
        \multirow{2}{*}{Def3DGS} & Ego & 29.75 $\pm$ 0.01 & 0.82 $\pm$ 0.001 & 0.27 $\pm$ 0.002 & 29.19 $\pm$ 0.04 & 0.81 $\pm$ 0.001 & 0.26 $\pm$ 0.000 \\
        & Exo & \cellcolor{moderategrey}33.95 $\pm$ 0.20 & \cellcolor{moderategrey}0.96 $\pm$ 0.004 & \cellcolor{moderategrey}0.12 $\pm$ 0.008 & \cellcolor{moderategrey}31.37 $\pm$ 0.17 & \cellcolor{moderategrey}0.94 $\pm$ 0.001 & \cellcolor{moderategrey}0.15 $\pm$ 0.001 \\
        \cmidrule{1-8}
        \multirow{2}{*}{4DGS} & Ego & 28.96 $\pm$ 0.02 & 0.79 $\pm$ 0.000 & 0.32 $\pm$ 0.000 & 29.35 $\pm$ 0.03 & 0.80 $\pm$ 0.001 & 0.29 $\pm$ 0.001 \\
        & Exo & \cellcolor{moderategrey}32.05 $\pm$ 0.05 & \cellcolor{moderategrey}0.91 $\pm$ 0.001 & \cellcolor{moderategrey}0.20 $\pm$ 0.001 & \cellcolor{moderategrey}30.14 $\pm$ 0.12 & \cellcolor{moderategrey}0.91 $\pm$ 0.002 & \cellcolor{moderategrey}0.20 $\pm$ 0.002 \\
        \cmidrule{1-8}
        \multirow{2}{*}{RTGS} & Ego & 29.27 $\pm$ 0.01 & 0.81 $\pm$ 0.001 & 0.30 $\pm$ 0.000 & \cellcolor{moderategrey}29.26 $\pm$ 0.01 & 0.81 $\pm$ 0.000 & 0.29 $\pm$ 0.001 \\
        & Exo & \cellcolor{moderategrey}31.30 $\pm$ 0.05 & \cellcolor{moderategrey}0.92 $\pm$ 0.003 & \cellcolor{moderategrey}0.16 $\pm$ 0.004 & 28.05 $\pm$ 0.17 & \cellcolor{moderategrey}0.88 $\pm$ 0.003 & \cellcolor{moderategrey}0.21 $\pm$ 0.009 \\
        \cmidrule[0.5pt]{1-8}
    \end{tabular}
    \caption{Performance of selected models on paired ego/exo views from 20 random scenes and 5 EgoGaussian-style scenes \cite{EgoGaussian}. Results are averaged over 3 runs ($\pm$ std). Grey indicates whether ego or exo scored higher. As we can observe, models perform noticeably better on exo views.}
    \label{tab:full_results}
\end{table}

%% file: tables/dynamic_results.tex
\begin{table*}[t!]
    \notsotiny
\begin{adjustwidth}{-2cm}{-2cm}
    \setlength{\tabcolsep}{3pt}
    \centering
    \begin{tabular}{cccccccccccccc}
        \cmidrule[0.5pt]{1-14}
        & & \multicolumn{6}{c}{\textbf{Dynamic masks}} & \multicolumn{6}{c}{\textbf{Static masks}} \\
         & & \multicolumn{3}{c}{\textbf{Random scenes}} & \multicolumn{3}{c}{\textbf{EgoGaussian scenes}} & \multicolumn{3}{c}{\textbf{Random scenes}} & \multicolumn{3}{c}{\textbf{EgoGaussian scenes}} \\
        \cmidrule(lr){3-5} \cmidrule(lr){6-8} \cmidrule(lr){9-11} \cmidrule(lr){12-14} 
        \textbf{Model} & \textbf{View} & mPSNR $\uparrow$ & mSSIM $\uparrow$ & mLPIPS $\downarrow$ & mPSNR $\uparrow$ & mSSIM $\uparrow$ & mLPIPS $\downarrow$ & mPSNR $\uparrow$ & mSSIM $\uparrow$ & mLPIPS $\downarrow$ & mPSNR $\uparrow$ & mSSIM $\uparrow$ & mLPIPS $\downarrow$\\
        \cmidrule{1-14}
        EgoGaussian & Ego & - & - & - & 23.03 $\pm$ 0.61 & 0.59 $\pm$ 0.027 & 0.38 $\pm$ 0.021 & - & - & - & 27.19 $\pm$ 0.04 & 0.77 $\pm$ 0.001 & 0.30 $\pm$ 0.001 \\
        \cmidrule{1-14}
        \multirow{2}{*}{Def3DGS} & Ego & \cellcolor{moderategrey}23.84 $\pm$ 0.04 & 0.64 $\pm$ 0.001 & 0.41 $\pm$ 0.000 & \cellcolor{moderategrey}24.99 $\pm$ 0.21 & 0.61 $\pm$ 0.010 & \cellcolor{moderategrey}0.36 $\pm$ 0.010 & 30.86 $\pm$ 0.02 & 0.83 $\pm$ 0.002 & 0.27 $\pm$ 0.002 & 29.38 $\pm$ 0.04 & 0.81 $\pm$ 0.001 & 0.26 $\pm$ 0.000\\
        & Exo & 22.17 $\pm$ 0.05 & \cellcolor{moderategrey}0.65 $\pm$ 0.004 & \cellcolor{moderategrey}0.31 $\pm$ 0.006 & 21.66 $\pm$ 0.09 & \cellcolor{moderategrey}0.63 $\pm$ 0.006 & 0.38 $\pm$ 0.006 & \cellcolor{moderategrey}37.23 $\pm$ 0.28 & \cellcolor{moderategrey}0.97 $\pm$ 0.004 & \cellcolor{moderategrey}0.11 $\pm$ 0.008 & \cellcolor{moderategrey}34.40 $\pm$ 0.21 & \cellcolor{moderategrey}0.96 $\pm$ 0.001 & \cellcolor{moderategrey}0.12 $\pm$ 0.001 \\
        \cmidrule{1-14}
        \multirow{2}{*}{4DGS} & Ego & \cellcolor{moderategrey}22.61 $\pm$ 0.03 & 0.59 $\pm$ 0.001 & 0.46 $\pm$ 0.001 & \cellcolor{moderategrey}24.96 $\pm$ 0.05 & 0.61 $\pm$ 0.004 & \cellcolor{moderategrey}0.39 $\pm$ 0.005 & 30.24 $\pm$ 0.03 & 0.80 $\pm$ 0.000 & 0.32 $\pm$ 0.000 & 29.56 $\pm$ 0.03 & 0.80 $\pm$ 0.001 & 0.29 $\pm$ 0.001 \\
        & Exo & 22.49 $\pm$ 0.05 & \cellcolor{moderategrey}0.66 $\pm$ 0.002 & \cellcolor{moderategrey}0.32 $\pm$ 0.002 & 21.95 $\pm$ 0.16 & \cellcolor{moderategrey}0.62 $\pm$ 0.004 & 0.40 $\pm$ 0.003 & \cellcolor{moderategrey}33.74 $\pm$ 0.06 & \cellcolor{moderategrey}0.92 $\pm$ 0.001 & \cellcolor{moderategrey}0.20 $\pm$ 0.001 & \cellcolor{moderategrey}32.09 $\pm$ 0.18 & \cellcolor{moderategrey}0.93 $\pm$ 0.002 & \cellcolor{moderategrey}0.18 $\pm$ 0.003 \\
        \cmidrule{1-14}
        \multirow{2}{*}{RTGS} & Ego & 23.52 $\pm$ 0.04 & 0.62 $\pm$ 0.001 & 0.42 $\pm$ 0.001 & \cellcolor{moderategrey}26.07 $\pm$ 0.04 & \cellcolor{moderategrey}0.65 $\pm$ 0.002 & \cellcolor{moderategrey}0.33 $\pm$ 0.002 & 30.34 $\pm$ 0.01 & 0.82 $\pm$ 0.001 & 0.30 $\pm$ 0.000 & \cellcolor{moderategrey}29.39 $\pm$ 0.02 & 0.81 $\pm$ 0.000 & 0.29 $\pm$ 0.001 \\
        & Exo & \cellcolor{moderategrey}23.62 $\pm$ 0.19 & \cellcolor{moderategrey}0.69 $\pm$ 0.003 & \cellcolor{moderategrey}0.26 $\pm$ 0.004 & 22.80 $\pm$ 0.08 & 0.64 $\pm$ 0.003 & 0.33 $\pm$ 0.003 & \cellcolor{moderategrey}32.47 $\pm$ 0.03 & \cellcolor{moderategrey}0.93 $\pm$ 0.003 & \cellcolor{moderategrey}0.16 $\pm$ 0.004 & 29.10 $\pm$ 0.25 & \cellcolor{moderategrey}0.90 $\pm$ 0.005 & \cellcolor{moderategrey}0.19 $\pm$ 0.010 \\
        \cmidrule[0.5pt]{1-14}
    \end{tabular}
\end{adjustwidth}
    \caption{Performance of selected models on dynamic and static masks for paired ego/exo views from 20 random scenes and 5 EgoGaussian-style scenes \cite{EgoGaussian}. Results are averaged over 3 runs ($\pm$ std). Grey indicates whether ego or exo view performs better. mPSNR is similar between ego/exo for dynamic masks; exo remains easier for static.}
    \label{tab:dynamic_results}
\end{table*}

%% file: tables/egogaussian_ek_no_sub_results.tex
\begin{table*}[t!]
    \scriptsize
    \setlength{\tabcolsep}{2pt}
    \centering
    \begin{tabular}{ccccccccccccc}
        \cmidrule[0.5pt]{1-13}
         & \multicolumn{6}{c}{\textbf{Epic-Kitchens}} & \multicolumn{6}{c}{\textbf{HOI4D}} \\
         & \multicolumn{3}{c}{\textbf{Passive segments}} & \multicolumn{3}{c}{\textbf{Active segments}} & \multicolumn{3}{c}{\textbf{Passive segments}} & \multicolumn{3}{c}{\textbf{Active segments}} \\
        \cmidrule(lr){2-4} \cmidrule(lr){5-7} \cmidrule(lr){8-10} \cmidrule(lr){11-13} 
        \textbf{Model} & mPSNR $\uparrow$ & mSSIM $\uparrow$ & mLPIPS $\downarrow$ & mPSNR $\uparrow$ & mSSIM $\uparrow$ & mLPIPS $\downarrow$ & mPSNR $\uparrow$ & mSSIM $\uparrow$ & mLPIPS $\downarrow$ & mPSNR $\uparrow$ & mSSIM $\uparrow$ & mLPIPS $\downarrow$\\
        \cmidrule{1-13}
        EgoGaussian (original) & 28.33 & 0.85 & 0.19 & 28.34 & 0.88 & 0.17 & 30.99 & 0.96 & 0.08 & 30.33 & 0.95 & 0.09 \\
        EgoGaussian (ours$^*$) & 28.76 & 0.86 & 0.18 & 30.55 & 0.89 & 0.15 & 30.52 & 0.96 & 0.09 & 31.12 & 0.96 & 0.09 \\
        EgoGaussian (ours$^\dagger$) & 28.61 & 0.85 & 0.25 & 30.35 & 0.88 & 0.22 & 30.43 & 0.95 & 0.13 & 30.97 & 0.95 & 0.13\\
        \cmidrule{1-13}
        Def3DGS (original) & 27.63 & 0.86 & 0.17 & 23.27 & 0.82 & 0.21 & 28.09 & 0.94 & 0.08 & 26.92 & 0.94 & 0.10 \\
        Def3DGS (ours$^\dagger$) & \cellcolor{moderategrey}37.54 & \cellcolor{moderategrey}0.96 & \cellcolor{moderategrey}0.12 & \cellcolor{moderategrey}32.94 & \cellcolor{moderategrey}0.94 & \cellcolor{moderategrey}0.16 & 34.38 & \cellcolor{moderategrey}0.97 & \cellcolor{moderategrey}0.08 & 33.06 & 0.96 & \cellcolor{moderategrey}0.10 \\
        \cmidrule{1-13}
        4DGS (original) & 28.90 & 0.87 & 0.16 & 23.13 & 0.80 & 0.23 & 28.69 & 0.94 & 0.08 & 27.33 & 0.94 & 0.10 \\
        4DGS (ours$^\dagger$) & 34.40 & 0.93 & 0.18 & 29.61 & 0.89 & 0.23 & \cellcolor{moderategrey}36.73 & \cellcolor{moderategrey}0.97 & 0.09 & \cellcolor{moderategrey}35.13 & \cellcolor{moderategrey}0.97 & 0.11\\
        \cmidrule[0.5pt]{1-13}
    \end{tabular}
    \caption{The performance of EgoGaussian and baselines on the original EgoGaussian Epic-Kitchens and HOI4D data \cite{epic_kitchens, hoi4d, EgoGaussian}. Ours$^*$ uses original metrics; ours$^\dagger$ uses corrected ones. Grey indicates the best score across ours$^\dagger$ runs. While ours$^*$ matches the original, baselines perform better than EgoGaussian, contrary to prior findings.}
    \label{tab:egogaussian_ek_no_sub_results.tex}
\end{table*}

%% file: figures/linear_velocity.tex
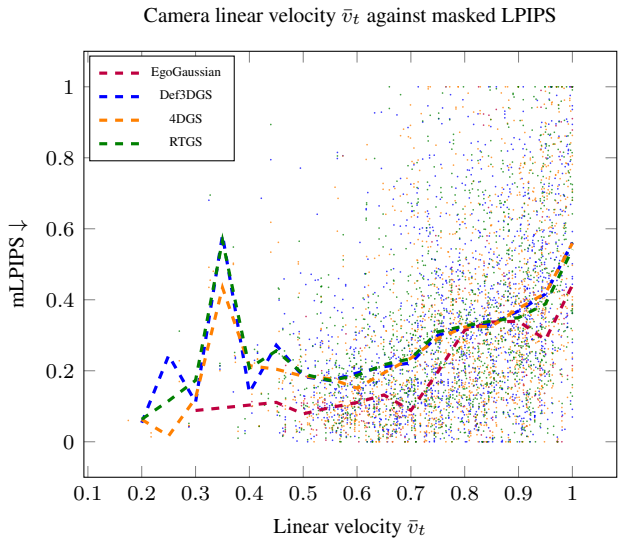
\begin{figure}[h]
    \centering
    \noindent
    \begin{tikzpicture}
    \begin{axis}[
        title=Camera linear velocity $\bar{v}_t$ against masked LPIPS,
        scale only axis,
        width=0.85\linewidth,
        height=0.85*0.8\linewidth,
        xlabel={Linear velocity $\bar{v}_t$},
        ylabel={mLPIPS $\downarrow$},
        xtick={0.1,0.2,0.3,0.4,0.5,0.6,0.7,0.8,0.9,1.0},
        ytick={0,0.2,0.4,0.6,0.8,1.0},
        legend style={at={(0.01,0.99)},anchor=north west, font=\tiny},
        tick label style={font=\footnotesize},
        label style={font=\footnotesize},
        title style={font=\footnotesize}
    ]
    
    \begin{pgfonlayer}{background layer}
    \foreach \model/\color in \modelcolormap {
        \edef\temp{\noexpand\addplot[
            color=\color,
            x=velocity,
            y=lpips,
            mark=*,
            only marks,
            mark size=0.1pt,
            opacity=0.8,
            forget plot
        ]
        table[col sep=comma] {figures/scatter_csvs/linear_scatter_data_\model.csv};
        }
        
        \temp
    }
    \end{pgfonlayer}
    
    \begin{pgfonlayer}{foreground layer}
    \foreach \model/\color in \modelcolormap {
        \edef\temp{
        \noexpand\addplot[
            color=\color,
            x=velocity,
            y=lpips,
            dashed,
            line width=1.2pt
        ]
        table[col sep=comma] {figures/trend_csvs/linear_trend_data_\model.csv};
        \noexpand\addlegendentry{\model}
        }
        
        \temp
    }
    \end{pgfonlayer}
    
    \end{axis}
    \end{tikzpicture}
    \caption{Camera linear velocity $\bar{v}_t$ plotted against mLPIPS. For legibility, only about 30\% of points are shown. Additional trend lines are computed and plotted based on all 100\% of points. As we can observe, as linear velocity increases, mLPIPS increases, which corresponds to worse reconstruction quality.}
    \label{fig:linear_velocity}
\end{figure}

%% file: figures/linear_baseline.tex
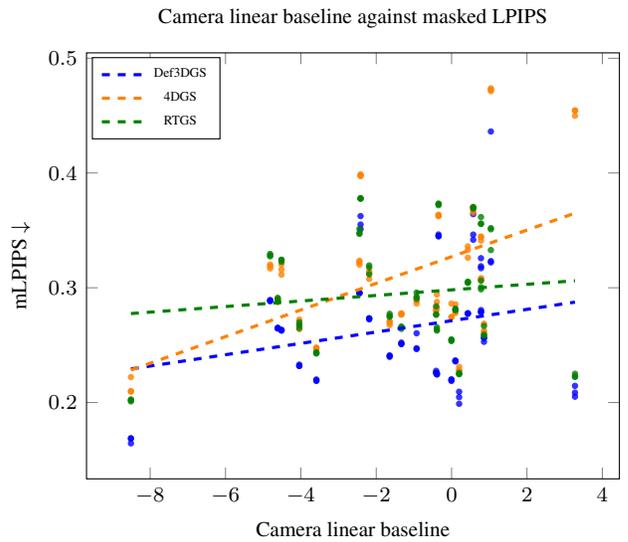
\begin{figure}[h]
    \centering
    \noindent
    \begin{tikzpicture}
    \begin{axis}[
        title=Camera linear baseline against masked LPIPS,
        scale only axis,
        width=0.85\linewidth,
        height=0.85*0.8\linewidth,
        xlabel={Camera linear baseline},
        ylabel={mLPIPS $\downarrow$},
        legend style={at={(0.01,0.99)},anchor=north west, font=\tiny},
        tick label style={font=\footnotesize},
        label style={font=\footnotesize},
        title style={font=\footnotesize}
    ]
    
    \begin{pgfonlayer}{background layer}
    \foreach \model/\color in \modelcolormapnoeg {
        \edef\temp{\noexpand\addplot[
            color=\color,
            x=baseline,
            y=lpips,
            mark=*,
            only marks,
            mark size=1.0pt,
            opacity=0.8,
            forget plot
        ]
        table[col sep=comma] {figures/baseline_csvs/linear_baseline_data_\model.csv};
        }
        
        \temp
    }
    \end{pgfonlayer}
    
    \begin{pgfonlayer}{foreground layer}
    \foreach \model/\color in \modelcolormapnoeg {
        \edef\temp{
        \noexpand\addplot[
            color=\color,
            x=baseline,
            y=lpips,
            line width=1.2pt,
            dashed
        ]
        table[col sep=comma] {figures/baseline_linear_csvs/linear_baseline_linear_data_\model.csv};
        \noexpand\addlegendentry{\model}
        }
        
        \temp
    }
    \end{pgfonlayer}
    
    \end{axis}
    \end{tikzpicture}
    \caption{Logarithmic camera linear baseline plotted against mLPIPS. Additional linear regression models fitted are shown. As we can see, no clear correlation can be observed in the graph.}
    \label{fig:linear_baseline}
\end{figure}

%% file: sec/5_discussion.tex
\section{Discussion}
\label{sec:discussion}

\textbf{Limitations.} One might question the fairness of comparing egocentric and exocentric recordings due to inherent differences in aspects such as camera motion and multi-view coverage. Egocentric video is recorded from a moving, first-person perspective, while exocentric footage relies on multiple static cameras positioned around the scene. However, these differences are not artifacts of the evaluation setup but fundamental properties of each modality. Fixed-camera, multi-view exocentric recordings are standard in ego-exo datasets \cite{ego_exo, assembly101, h2o, home_action_genome, cmu_mmac, ego_exo_learn, castle}, and reflect how such data is typically collected in practice. Therefore, while the two modalities differ substantially, comparing them still offers meaningful insights into model performance under realistic and representative conditions.

Due to the static exo cameras, a limitation of our current setup is that the exocentric camera poses used for testing are also seen during training, but at different timesteps. This allows the models to potentially memorize specific viewpoints, rather than generalize to novel views, effectively reducing the difficulty of the exo task. In contrast, the egocentric setting naturally enforces both temporal and spatial generalization due to the moving camera. While this introduces an asymmetry in the evaluation, it reflects a practical constraint of working with existing ego-exo datasets, where only a limited number of static cameras is available~\cite{ego_exo, charades_ego, assembly101, h2o, home_action_genome, cmu_mmac, castle, ego_exo_learn}. Future datasets could address this issue through denser exo coverage or the use of moving exocentric cameras.

Our evaluation is limited to 25 scenes due to the computational cost of training dynamic 3D Gaussian Splatting models, which often requires several hours of optimization per scene per method. However, this evaluation scale already exceeds prior datasets used for the evaluation of dynamic monocular 3DGS models such as D-NeRF (8 scenes) \cite{d_nerf}, Nerfies (4 scenes) \cite{nerfies}, HyperNeRF (17 scenes) \cite{hyper_nerf}, NeRF-DS (7 scenes) \cite{nerf_ds}, and DyCheck (14 scenes) \cite{dycheck}. While a limited number of scenes might make the reported numbers sensitive to outliers, we address this in Supplementary B by including per-scene comparisons between egocentric and exocentric views, showing that our main conclusions hold consistently across scenes.

\textbf{Egocentric difficulty.} Results from \Cref{tab:full_results} suggest that reconstruction from the egocentric perspective is more challenging for existing models than from the corresponding exocentric views. Interestingly, as shown in \Cref{tab:dynamic_results}, the difficulty of reconstructing from the ego perspective appears to stem from the static regions of the scene, with dynamic regions reconstructed at comparable mean mPSNR across both modalities. While these findings should be interpreted in light of the evaluation asymmetry, where exocentric test views coincide with training camera poses, this trend is further reinforced by the camera motion results. 

The camera motion results suggest a negative correlation between camera motion in egocentric videos and reconstruction quality. This contrasts with prior claims in non-egocentric settings \cite{monocular_evaluation, dycheck}. Hence, these observations reinforce the finding that reconstruction from an egocentric perspective presents distinct challenges compared to other types of data. Importantly, camera motion is unlikely to be the sole factor affecting quality. Its correlation with reconstruction performance may also reflect underlying influences such as body movement.

\textbf{Dynamic difficulty.} The results from \Cref{tab:dynamic_results} show that the reconstruction of dynamics is worse than the reconstruction of the static background. Existing methods reconstruct the static background with a higher visual fidelity than the dynamic objects. These results may appear surprising, as in \cite{monocular_evaluation}, the authors found that the performance of dynamic Gaussian methods `does not change much` after masking out the static components. Our results thus show that it is still worthwhile to evaluate static and dynamic regions separately when developing future models. Likewise, it might be beneficial to model static and dynamic regions separately, such as in \cite{gaufre} or \cite{EgoGaussian}.

% Another limitation is reporting metrics in terms of mean value over all scenes. Since the performance may vary greatly between scenes due to their difficulty, outliers may skew the means. In \Cref{sec:per_scene_results}, we address this problem by comparing ego and exo performance per-scene and showing that the same conclusions hold.

% Finally, we note that the ego-exo analysis included only two EgoGaussian scenes. This limitation stems primarily from time constraints and the lengthy training time required for EgoGaussian. Moreover, as our findings also suggest, the constraints posed by EgoGaussian make the scenes less representative of real-world.

\textbf{Future work.} Our results suggest that egocentric reconstruction is more challenging than exocentric, but due to limited static exo cameras, spatial generalization in exocentric evaluation is restricted. This asymmetry means the ego-exo comparison should be interpreted cautiously. Still, reinforced by observed negative correlations of egocentric camera motion to reconstruction quality, the findings highlight the need for egocentric-specific models. Future models should also not neglect the reconstruction of static regions, as they may contribute to the difference between ego and exo reconstruction. Contrary to prior work \cite{monocular_evaluation}, we find that dynamic regions are reconstructed less accurately than static ones, revealing a key blind spot in current models. Future benchmarks should therefore separate static and dynamic evaluation. Finally, future datasets with denser or moving exocentric cameras are needed to enable fairer ego-exo comparisons and drive improved methods.

%% file: sec/6_conclusion.tex
\section{Conclusion}
\label{sec:conclusion}

In this work, we answered the question of how well existing, monocular dynamic 3D Gaussian Splatting models perform in egocentric settings. To this end, we compared the performance of existing models on paired ego-exo views from the EgoExo4D dataset. We found that models tend to achieve better reconstruction quality of scenes captured from the exocentric perspective. Additionally, our results suggest that this difference, measured in masked peak signal-to-noise ratio (PSNR), comes from the reconstruction of static parts of the scene, as the dynamic regions tend to be reconstructed with similar PSNR quality between ego and exo views. Overall, our results show the need for models specialized in egocentric reconstruction, as current models struggle with the challenges posed by egocentric video.

%% file: sec/X_suppl.tex
\clearpage
\appendix
\setcounter{page}{1}
\maketitlesupplementary

\section{Hyperparameter search space}
\label{sec:hyper_search_space}

The search spaces of hyperparameters for Deformable-3DGS, RTGS, and 4DGS are shown in Tables \ref{tab:hyper_def3dgs}, \ref{tab:hyper_rtgs}, and \ref{tab:hyper_4dgs} respectively. For 4DGS and RTGS these search spaces were obtained by combining all existing configurations. For Deformable-3DGS, we search over the shape of the deformation field as well as the total number of iterations. 

\input{tables/hyper_def3dgs}
\input{tables/hyper_rtgs}
\input{tables/hyper_4dgs}

\section{Per-scene results}
\label{sec:per_scene_results}

The results in the main paper represent the metrics averaged over scenes and over 3 runs. Since the scenes can vary greatly in difficulty, outliers may skew the metrics. To this end, in this section, we evaluate the models per-scene. For each model and metric, we count in how many runs the exo performance was better than ego and we report the ratio of this number to the total number of runs. The total number of runs is 60 for random scenes and 15 for EgoGaussian scenes. 

\textbf{Full results.} The results without the static-dynamic separation are presented in \Cref{tab:per_scene_full}. As we can see for the random scenes, all ratios are above 0.5, indicating that most scenes were reconstructed better from the exo view, which reinforces previous findings. In the EgoGaussian scenes, mSSIM and mLPIPS are always better on the exo views while mPSNR is better in only some of the runs. This corresponds to the previous results, where we observed the mPSNR to be closer between ego and exo on the EgoGaussian scenes.

\textbf{Dynamic and static results.} The comparison on the dynamic masks is shown in \Cref{tab:per_scene_dynamic}. As we can see, the mPSNR ratio has visibly dropped to either below 0.5 or close to 0.5, which is in line with the previous findings, where the mean mPSNR was either higher in the ego view or similar. In the static reconstruction from \Cref{tab:per_scene_static} we again observe that the static mPSNR is higher in more random scenes on the exo view than the ego. Therefore, these results also suggest that the mPSNR-performance between ego and exo views stems primarily from the reconstruction of static objects. Since for mSSIM and mLPIPS the exo reconstruction is also easier on the dynamic masks, no such conclusion can be made for these metrics. These results therefore reinforce the findings from the main paper.

\input{tables/suppl/per_scene_full}
\input{tables/suppl/per_scene_dynamic}
\input{tables/suppl/per_scene_static}

\section{Dataset}
\label{sec:dataset_full}

10 example scenes included in our dataset have been shown in \Cref{fig:dataset_full}. The full dataset will be released in the future.

\input{figures/dataset_full}

\section{Extra results for camera motion}
\label{sec:camera_motion_extra}

\subsection{Camera velocity}

\textbf{Results for angular velocity.} Let $\bar{\omega}_t$ be the angular velocity of the camera, normalized identically to $\bar{v}_t$. \Cref{fig:angular_velocity} then shows the relationship between $\bar{\omega}_t$ and the reconstruction quality measured in terms of mLPIPS. As we can observe, as angular camera motion increases, reconstruction worsens. This is in line with the results for linear velocity.

\textbf{Results for other metrics.} \Cref{fig:velocity_psnr_ssim} presents the results for linear and angular velocities when measuring mPSNR and mSSIM. As we can observe, both metrics tend to decrease. This again shows worsening reconstruction quality and hence reinforces previous results.

\input{figures/suppl/linear_angular_velocities_psnr_ssim}

\textbf{Significance test results.} \Cref{tab:linear_velocity_significance} presents the Pearson and Spearman coefficient results for linear velocity on all 3 metrics. As we can observe, all coefficients coincide with a negative correlation between camera velocity and reconstruction quality. Additionally, all p-values are far below 0.05, indicating statistical significance of results. Similar results can be seen for angular velocity in \Cref{tab:angular_velocity_significance}.

\input{tables/suppl/velocity_significance}

\subsection{Camera baseline}

\textbf{Results for angular baseline.} Similarly to the linear baseline from the main paper, we can define the angular baseline to be the highest angular difference between any two camera poses. \Cref{fig:angular_baseline} presents the camera angular baseline plotted against mLPIPS. As we can observe, an increase in camera baseline does not clearly correlate with worse reconstruction quality, similarly to the linear baseline from the main paper.

\textbf{Results for other metrics.} \Cref{fig:baseline_psnr_ssim} presents the results for linear and angular baselines when measuring mPSNR and mSSIM. Interestingly, for mPSNR a clear decreasing pattern can be observed, which cannot be observed for mSSIM or mLPIPS. Importantly, however, we did not observe a pattern indicating that increasing the camera baseline would lead to better reconstruction. This is in line with findings from the main paper.

\input{figures/suppl/linear_angular_baselines_psnr_ssim}

\textbf{Significance test results.} \Cref{tab:linear_baseline_significance} presents the Pearson and Spearman coefficient results for linear baseline on all 3 metrics. Although not all results are statistically significant ($p \geq 0.05$), all coefficients indicate a negative correlation between baseline and reconstruction quality, and no positive correlation has been observed. Similar results can be seen for angular baseline in \Cref{tab:angular_baseline_significance}. These results reinforce our findings that increasing camera motion, expressed in terms of baseline, does not correlate with an increase in reconstruction quality, and may in fact instead hinder it. 

\input{tables/suppl/baseline_signficance}

\input{figures/angular_velocity}
\input{figures/angular_baseline}

%% file: tables/hyper_def3dgs.tex
\begin{table}[h]
    \centering
    \begin{tabular}{c|c}
        \textbf{Parameter name} & \textbf{Values} \\
        \hline
        \texttt{iterations} & $1.5 \cdot 10^{4}$, $2.5 \cdot 10^{4}$, $4.0 \cdot 10^{4}$ \\
        \texttt{deform\_depth} & $6$, $8$, $10$ \\
        \texttt{deform\_width} & $128$, $256$, $512$ \\
    \end{tabular}
    \caption{The hyperparameter search space for Deformable-3DGS.}
    \label{tab:hyper_def3dgs}
\end{table}

%% file: tables/hyper_rtgs.tex
\begin{table}[h]
    \centering
    \begin{tabular}{c|c}
        \textbf{Parameter name} & \textbf{Values} \\
        \hline
        \texttt{env\_map\_res} & $0$, $500$ \\
        \texttt{env\_optimize\_until} & $1.0 \cdot 10^{9}$, $5.0 \cdot 10^{3}$ \\
        \texttt{iterations} & $2.0 \cdot 10^{4}$, $3.0 \cdot 10^{4}$ \\
        \texttt{position\_lr\_max\_steps} & $1.5 \cdot 10^{4}$, $3.0 \cdot 10^{4}$ \\
        \texttt{densification\_interval} & $200$, $100$ \\
        \texttt{densify\_until\_iter} & $1.0 \cdot 10^{4}$, $1.5 \cdot 10^{4}$ \\
    \end{tabular}
    \caption{The hyperparameter search space for RTGS.}
    \label{tab:hyper_rtgs}
\end{table}

%% file: tables/hyper_4dgs.tex
\begin{table*}[t]
    \centering
    \begin{tabular}{c|c}
        \textbf{Parameter name} & \textbf{Values} \\
        \hline
        \texttt{grid\_dimensions} & $2$ \\
        \texttt{input\_coordinate\_dim} & $4$ \\
        \texttt{output\_coordinate\_dim} & $16$, $32$ \\
        \texttt{resolution[-1]} & $250$, $150$, $100$, $80$, $75$, $50$, $25$ \\
        \texttt{multires} & [$1$, $2$, $4$], [$1$, $2$] \\
        \texttt{defor\_depth} & $1$, $0$ \\
        \texttt{net\_width} & $128$, $64$ \\
        \texttt{plane\_tv\_weight} & $2.0 \cdot 10^{-4}$, $1.0 \cdot 10^{-4}$ \\
        \texttt{time\_smoothness\_weight} & $1.0 \cdot 10^{-3}$, $1.0 \cdot 10^{-2}$ \\
        \texttt{l1\_time\_planes} & $1.0 \cdot 10^{-4}$ \\
        \texttt{no\_do} & True, False \\
        \texttt{no\_dshs} & True, False \\
        \texttt{no\_ds} & True, False \\
        \texttt{iterations} & $1.4 \cdot 10^{4}$, $1.5 \cdot 10^{4}$, $2.0 \cdot 10^{4}$ \\
        \texttt{batch\_size} & $1$, $2$ \\
        \texttt{coarse\_iterations} & $3.0 \cdot 10^{3}$ \\
        \texttt{densify\_until\_iter} & $1.0 \cdot 10^{4}$, $1.5 \cdot 10^{4}$ \\
        \texttt{opacity\_reset\_interval} & $3.0 \cdot 10^{3}$, $3.0 \cdot 10^{6}$ \\
        \texttt{grid\_lr\_init} & $1.6 \cdot 10^{-3}$ \\
        \texttt{grid\_lr\_final} & $1.6 \cdot 10^{-4}$, $1.6 \cdot 10^{-5}$ \\
        \texttt{opacity\_threshold\_coarse} & $5.0 \cdot 10^{-3}$ \\
        \texttt{opacity\_threshold\_fine\_init} & $5.0 \cdot 10^{-3}$ \\
        \texttt{opacity\_threshold\_fine\_after} & $5.0 \cdot 10^{-3}$ \\
        \texttt{pruning\_interval} & $100$, $8.0 \cdot 10^{3}$ \\
        \texttt{deformation\_lr\_init} & $1.6 \cdot 10^{-4}$ \\
        \texttt{deformation\_lr\_final} & $1.6 \cdot 10^{-5}$, $1.6 \cdot 10^{-6}$ \\
        \texttt{deformation\_lr\_delay\_mult} & $1.0 \cdot 10^{-2}$ \\
    \end{tabular}
    \caption{The hyperparameter search space for 4DGS.}
    \label{tab:hyper_4dgs}
\end{table*}

%% file: tables/suppl/per_scene_full.tex
\begin{table*}[h]
    \centering
    \begin{tabular}{ccccccc}
        \cmidrule[0.5pt]{1-7}
        & \multicolumn{3}{c}{\textbf{Random scenes}} & \multicolumn{3}{c}{\textbf{EgoGaussian scenes}} \\
        \cmidrule(lr){2-4} \cmidrule(lr){5-7} 
        \textbf{Model} & mPSNR & mSSIM & mLPIPS & mPSNR & mSSIM & mLPIPS \\
        \cmidrule{1-7}
        Deformable-3DGS & 0.98 & 0.98 & 0.98 & 0.73 & 1.00 & 1.00 \\
        4DGS & 0.80 & 0.95 & 0.85 & 0.60 & 1.00 & 1.00 \\
        RTGS & 0.77 & 0.92 & 0.93 & 0.20 & 1.00 & 1.00 \\
        \cmidrule[0.5pt]{1-7}
    \end{tabular}
    \caption{Per-scene results of the baselines on 20 random and 5 EgoGaussian scenes with 3 runs per scene. Each entry represents the ratio of runs where the performance on the exo view was higher than the corresponding ego view to the total number of runs.}
    \label{tab:per_scene_full}
\end{table*}

%% file: tables/suppl/per_scene_dynamic.tex
\begin{table*}[h]
    \centering
    \begin{tabular}{ccccccc}
        \cmidrule[0.5pt]{1-7}
        & \multicolumn{3}{c}{\textbf{Random scenes}} & \multicolumn{3}{c}{\textbf{EgoGaussian scenes}} \\
        \cmidrule(lr){2-4} \cmidrule(lr){5-7} 
        \textbf{Model} & mPSNR & mSSIM & mLPIPS & mPSNR & mSSIM & mLPIPS \\
        \cmidrule{1-7}
        Deformable-3DGS & 0.33 & 0.62 & 0.82 & 0.20 & 0.60 & 0.47 \\
        4DGS & 0.59 & 0.78 & 0.85 & 0.20 & 0.67 & 0.27 \\
        RTGS & 0.52 & 0.80 & 0.93 & 0.20 & 0.40 & 0.33 \\
        \cmidrule[0.5pt]{1-7}
    \end{tabular}
    \caption{Per-scene results of the baselines on 20 random and 5 EgoGaussian scenes with 3 runs per scene. Dynamic mask considered only. Each entry represents the ratio of runs where the performance on the exo view was higher than the corresponding ego view to the total number of runs.}
    \label{tab:per_scene_dynamic}
\end{table*}

%% file: tables/suppl/per_scene_static.tex
\begin{table*}[h]
    \centering
    \begin{tabular}{ccccccc}
        \cmidrule[0.5pt]{1-7}
        & \multicolumn{3}{c}{\textbf{Random scenes}} & \multicolumn{3}{c}{\textbf{EgoGaussian scenes}} \\
        \cmidrule(lr){2-4} \cmidrule(lr){5-7} 
        \textbf{Model} & mPSNR & mSSIM & mLPIPS & mPSNR & mSSIM & mLPIPS \\
        \cmidrule{1-7}
        Deformable-3DGS & 0.98 & 0.98 & 0.98 & 1.00 & 1.00 & 1.00 \\
        4DGS & 0.85 & 0.95 & 0.85 & 0.87 & 1.00 & 1.00 \\
        RTGS & 0.83 & 0.92 & 0.92 & 0.40 & 1.00 & 1.00 \\
        \cmidrule[0.5pt]{1-7}
    \end{tabular}
    \caption{Per-scene results of the baselines on 20 random and 5 EgoGaussian scenes with 3 runs per scene. Static mask considered only. Each entry represents the ratio of runs where the performance on the exo view was higher than the corresponding ego view to the total number of runs.}
    \label{tab:per_scene_static}
\end{table*}

%% file: figures/dataset_full.tex
\begin{figure*}[t!]
    \centering
    \includegraphics[width=0.9\linewidth]{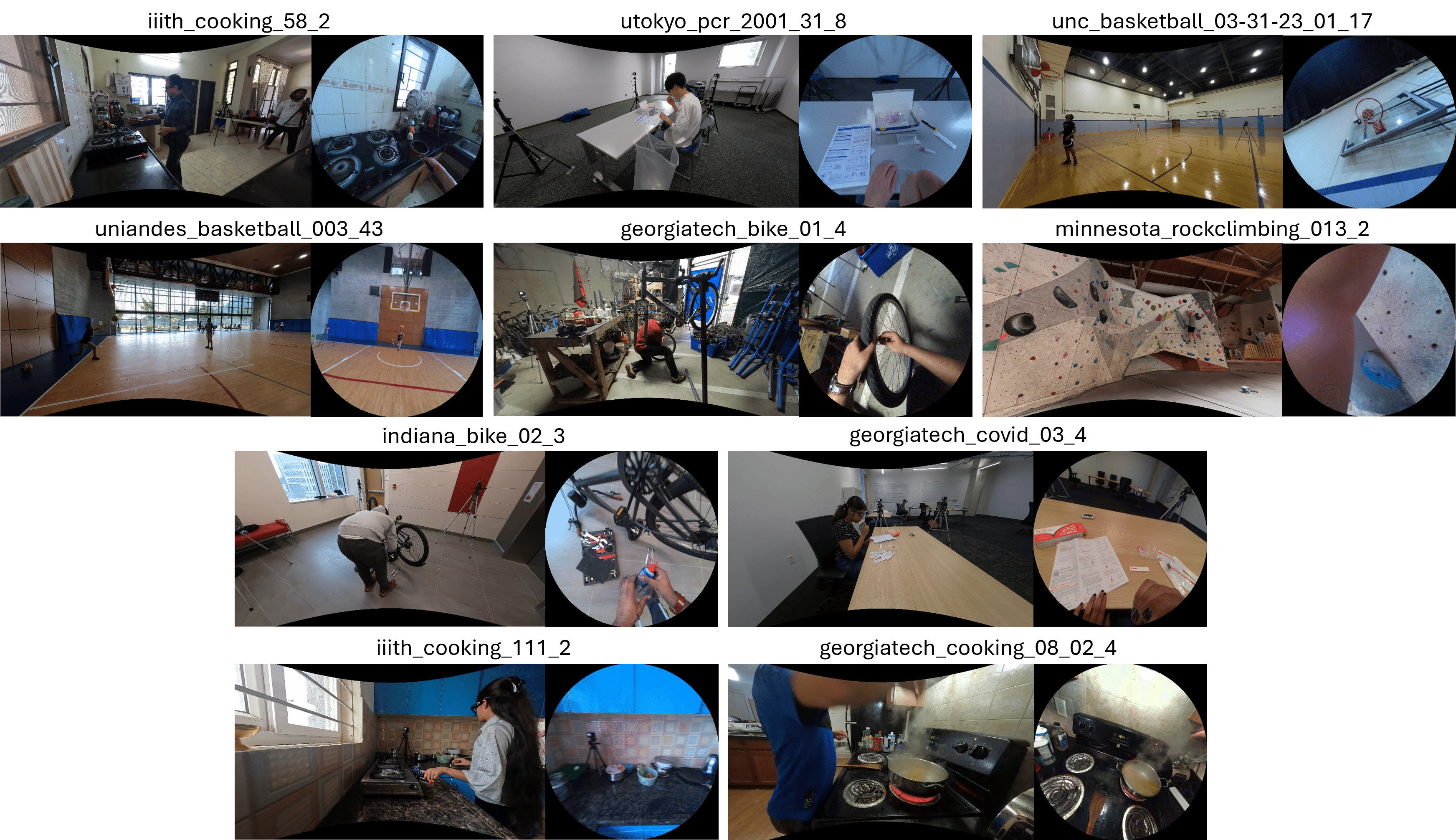}
    \caption{First frame of 10 example scenes from the exo (left, first exo camera) and ego (right) views. First 8 scenes are random; last 2 (bottom) are selected EgoGaussian-style scenes. Frames shown after undistortion. As we can see, the scenes are varied. Best viewed zoomed in.}
    \label{fig:dataset_full}
\end{figure*}

%% file: figures/suppl/linear_angular_velocities_psnr_ssim.tex
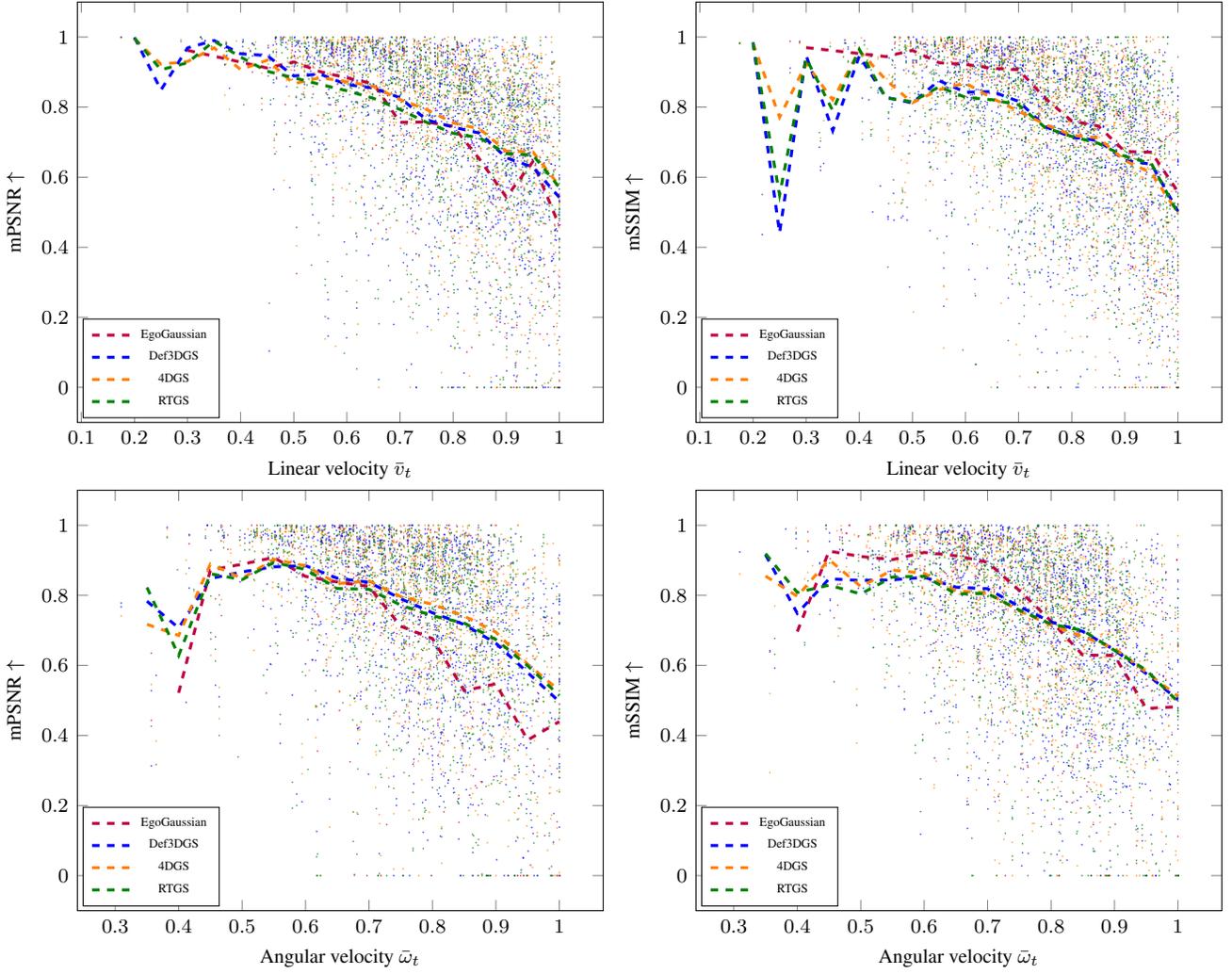
\begin{figure*}[t]
    \centering
    \begin{subfigure}[b]{0.5\textwidth}
        \centering
        \noindent
        \begin{tikzpicture}
        \begin{axis}[
            scale only axis,
            width=0.85\linewidth,
            height=0.85*0.8\linewidth,
            xlabel={Linear velocity $\bar{v}_t$},
            ylabel={mPSNR $\uparrow$},
            xtick={0.1,0.2,0.3,0.4,0.5,0.6,0.7,0.8,0.9,1.0},
            ytick={0,0.2,0.4,0.6,0.8,1.0},
            legend style={at={(0.01,0.01)},anchor=south west, font=\tiny},
            tick label style={font=\footnotesize},
            label style={font=\footnotesize}
        ]
        
        \begin{pgfonlayer}{background layer}
        \foreach \model/\color in \modelcolormap {
            \edef\temp{\noexpand\addplot[
                color=\color,
                x=velocity,
                y=lpips,
                mark=*,
                only marks,
                mark size=0.1pt,
                opacity=0.8,
                forget plot
            ]
            table[col sep=comma] {figures/suppl/psnr_scatter_csvs/linear_scatter_data_\model.csv};
            }
            
            \temp
        }
        \end{pgfonlayer}
        
        \begin{pgfonlayer}{foreground layer}
        \foreach \model/\color in \modelcolormap {
            \edef\temp{
            \noexpand\addplot[
                color=\color,
                x=velocity,
                y=lpips,
                dashed,
                line width=1.2pt
            ]
            table[col sep=comma] {figures/suppl/psnr_trend_csvs/linear_trend_data_\model.csv};
            \noexpand\addlegendentry{\model}
            }
            
            \temp
        }
        \end{pgfonlayer}
        
        \end{axis}
        \end{tikzpicture}
    \end{subfigure}%
    \begin{subfigure}[b]{0.5\textwidth}
        \centering
        \noindent
        \begin{tikzpicture}
        \begin{axis}[
            scale only axis,
            width=0.85\linewidth,
            height=0.85*0.8\linewidth,
            xlabel={Linear velocity $\bar{v}_t$},
            ylabel={mSSIM $\uparrow$},
            xtick={0.1,0.2,0.3,0.4,0.5,0.6,0.7,0.8,0.9,1.0},
            ytick={0,0.2,0.4,0.6,0.8,1.0},
            legend style={at={(0.01,0.01)},anchor=south west, font=\tiny},
            tick label style={font=\footnotesize},
            label style={font=\footnotesize}
        ]
        
        \begin{pgfonlayer}{background layer}
        \foreach \model/\color in \modelcolormap {
            \edef\temp{\noexpand\addplot[
                color=\color,
                x=velocity,
                y=lpips,
                mark=*,
                only marks,
                mark size=0.1pt,
                opacity=0.8,
                forget plot
            ]
            table[col sep=comma] {figures/suppl/ssim_scatter_csvs/linear_scatter_data_\model.csv};
            }
            
            \temp
        }
        \end{pgfonlayer}
        
        \begin{pgfonlayer}{foreground layer}
        \foreach \model/\color in \modelcolormap {
            \edef\temp{
            \noexpand\addplot[
                color=\color,
                x=velocity,
                y=lpips,
                dashed,
                line width=1.2pt
            ]
            table[col sep=comma] {figures/suppl/ssim_trend_csvs/linear_trend_data_\model.csv};
            \noexpand\addlegendentry{\model}
            }
            
            \temp
        }
        \end{pgfonlayer}
        
        \end{axis}
        \end{tikzpicture}
    \end{subfigure} \\
    \begin{subfigure}[b]{0.5\textwidth}
        \centering
        \noindent
        \begin{tikzpicture}
        \begin{axis}[
            scale only axis,
            width=0.85\linewidth,
            height=0.85*0.8\linewidth,
            xlabel={Angular velocity $\bar{\omega}_t$},
            ylabel={mPSNR $\uparrow$},
            xtick={0.3,0.4,0.5,0.6,0.7,0.8,0.9,1.0},
            ytick={0,0.2,0.4,0.6,0.8,1.0},
            legend style={at={(0.01,0.01)},anchor=south west, font=\tiny},
            tick label style={font=\footnotesize},
            label style={font=\footnotesize}
        ]
        
        \begin{pgfonlayer}{background layer}
        \foreach \model/\color in \modelcolormap {
            \edef\temp{\noexpand\addplot[
                color=\color,
                x=velocity,
                y=lpips,
                mark=*,
                only marks,
                mark size=0.1pt,
                opacity=0.8,
                forget plot
            ]
            table[col sep=comma] {figures/suppl/psnr_scatter_csvs/angular_scatter_data_\model.csv};
            }
            
            \temp
        }
        \end{pgfonlayer}
        
        \begin{pgfonlayer}{foreground layer}
        \foreach \model/\color in \modelcolormap {
            \edef\temp{
            \noexpand\addplot[
                color=\color,
                x=velocity,
                y=lpips,
                dashed,
                line width=1.2pt
            ]
            table[col sep=comma] {figures/suppl/psnr_trend_csvs/angular_trend_data_\model.csv};
            \noexpand\addlegendentry{\model}
            }
            
            \temp
        }
        \end{pgfonlayer}
        
        \end{axis}
        \end{tikzpicture}
    \end{subfigure}%
    \begin{subfigure}[b]{0.5\textwidth}
        \centering
        \noindent
        \begin{tikzpicture}
        \begin{axis}[
            scale only axis,
            width=0.85\linewidth,
            height=0.85*0.8\linewidth,
            xlabel={Angular velocity $\bar{\omega}_t$},
            ylabel={mSSIM $\uparrow$},
            xtick={0.3,0.4,0.5,0.6,0.7,0.8,0.9,1.0},
            ytick={0,0.2,0.4,0.6,0.8,1.0},
            legend style={at={(0.01,0.01)},anchor=south west, font=\tiny},
            tick label style={font=\footnotesize},
            label style={font=\footnotesize}
        ]
        
        \begin{pgfonlayer}{background layer}
        \foreach \model/\color in \modelcolormap {
            \edef\temp{\noexpand\addplot[
                color=\color,
                x=velocity,
                y=lpips,
                mark=*,
                only marks,
                mark size=0.1pt,
                opacity=0.8,
                forget plot
            ]
            table[col sep=comma] {figures/suppl/ssim_scatter_csvs/angular_scatter_data_\model.csv};
            }
            
            \temp
        }
        \end{pgfonlayer}
        
        \begin{pgfonlayer}{foreground layer}
        \foreach \model/\color in \modelcolormap {
            \edef\temp{
            \noexpand\addplot[
                color=\color,
                x=velocity,
                y=lpips,
                dashed,
                line width=1.2pt
            ]
            table[col sep=comma] {figures/suppl/ssim_trend_csvs/angular_trend_data_\model.csv};
            \noexpand\addlegendentry{\model}
            }
            
            \temp
        }
        \end{pgfonlayer}
        
        \end{axis}
        \end{tikzpicture}
    \end{subfigure}
    \caption{Camera linear and angular velocity results for mPSNR and mSSIM. As we can see, increasing either velocity correlates with a decrease in metrics, indicating worse reconstruction performance.}
    \label{fig:velocity_psnr_ssim}
\end{figure*}

%% file: tables/suppl/velocity_significance.tex
\begin{table*}[t]
    \centering
    \begin{tabular}{cccccccc}
        \cmidrule[0.5pt]{1-8}
        & & \multicolumn{2}{c}{\textbf{mPSNR} $\uparrow$} & \multicolumn{2}{c}{\textbf{mSSIM} $\uparrow$} & \multicolumn{2}{c}{\textbf{mLPIPS} $\downarrow$} \\
        \cmidrule(lr){3-4} \cmidrule(lr){5-6} \cmidrule(lr){7-8} 
        \textbf{Model} & \textbf{Coefficient} & Value & p-value & Value & p-value & Value & p-value \\
        \cmidrule{1-8}
        \multirow{2}{*}{EgoGaussian} & Pearson & $-0.54$ & $5.3 \cdot 10^{-56}$ & $-0.52$ & $1.6 \cdot 10^{-50}$ & $0.47$ & $3.1 \cdot 10^{-40}$ \\
        & Spearman & $-0.56$ & $7.6 \cdot 10^{-61}$ & $-0.58$ & $1.2 \cdot 10^{-64}$ & $0.51$ & $1.3 \cdot 10^{-47}$ \\
        \cmidrule{1-8}
        \multirow{2}{*}{Def3DGS} & Pearson & $-0.44$ & $2.5 \cdot 10^{-248}$ & $-0.41$ & $5.0 \cdot 10^{-207}$ & $0.39$ & $2.6 \cdot 10^{-185}$\\
        & Spearman & $-0.50$ & $<2.2 \cdot 10^{-308}$ & $-0.46$ & $1.9 \cdot 10^{-265}$ & $0.41$ & $2.2 \cdot 10^{-214}$ \\
        \cmidrule{1-8}
        \multirow{2}{*}{4DGS} & Pearson & $-0.39$ & $1.3 \cdot 10^{-191}$ & $-0.42$ & $1.9 \cdot 10^{-218}$ & $0.40$ & $9.0 \cdot 10^{-198}$ \\
        & Spearman & $-0.43$ & $2.2 \cdot 10^{-239}$ & $-0.45$ & $1.0 \cdot 10^{-253}$ & $0.43$ & $2.0 \cdot 10^{-239}$ \\
        \cmidrule{1-8}
        \multirow{2}{*}{RTGS} & Pearson & $-0.37$ & $2.7 \cdot 10^{-164}$ & $-0.39$ & $1.2 \cdot 10^{-189}$ & $0.36$ & $1.1 \cdot 10^{-161}$ \\
        & Spearman & $-0.42$ & $4.3 \cdot 10^{-223}$ & $-0.41$ & $5.6 \cdot 10^{-206}$ & $0.38$ & $1.1 \cdot 10^{-176}$ \\
        \cmidrule[0.5pt]{1-8}
    \end{tabular}
    \caption{Significance test results for the linear velocity experiments. The p-value approximately indicates the probability that the correlation coefficient is 0. As we can observe, as linear velocity increases, the reconstruction quality decreases with p-values of far below 0.05.}
    \label{tab:linear_velocity_significance}
\end{table*}

\begin{table*}[t]
    \centering
    \begin{tabular}{cccccccc}
        \cmidrule[0.5pt]{1-8}
        & & \multicolumn{2}{c}{\textbf{mPSNR} $\uparrow$} & \multicolumn{2}{c}{\textbf{mSSIM} $\uparrow$} & \multicolumn{2}{c}{\textbf{mLPIPS} $\downarrow$} \\
        \cmidrule(lr){3-4} \cmidrule(lr){5-6} \cmidrule(lr){7-8} 
        \textbf{Model} & \textbf{Coefficient} & Value & p-value & Value & p-value & Value & p-value \\
        \cmidrule{1-8}
        \multirow{2}{*}{EgoGaussian} & Pearson & $-0.56$ & $6.0 \cdot 10^{-61}$ & $-0.58$ & $1.3 \cdot 10^{-63}$ & $0.59$ & $4.9 \cdot 10^{-67}$ \\ 
        & Spearman & $-0.57$ & $5.2 \cdot 10^{-62}$ & $-0.60$ & $3.1 \cdot 10^{-69}$ & $0.52$ & $6.1 \cdot 10^{-50}$ \\
        \cmidrule{1-8}
        \multirow{2}{*}{Def3DGS} & Pearson & $-0.42$ & $8.8 \cdot 10^{-221}$ & $-0.39$ & $5.2 \cdot 10^{-189}$ & $0.41$ & $7.1 \cdot 10^{-216}$ \\
        & Spearman & $-0.46$ & $7.9 \cdot 10^{-268}$ & $-0.42$ & $2.3 \cdot 10^{-225}$ & $0.43$ & $1.4 \cdot 10^{-232}$ \\
        \cmidrule{1-8}
        \multirow{2}{*}{4DGS} & Pearson & $-0.38$ & $4.0 \cdot 10^{-176}$ & $-0.38$ & $6.9 \cdot 10^{-175}$ & $0.42$ & $1.8 \cdot 10^{-219}$ \\
        & Spearman & $-0.41$ & $4.3 \cdot 10^{-215}$ & $-0.40$ & $3.3 \cdot 10^{-198}$ & $0.43$ & $2.7 \cdot 10^{-234}$ \\
        \cmidrule{1-8}
        \multirow{2}{*}{RTGS} & Pearson & $-0.37$ & $4.8 \cdot 10^{-165}$ & $-0.37$ & $2.3 \cdot 10^{-169}$ & $0.42$ & $2.1 \cdot 10^{-224}$ \\
        & Spearman & $-0.42$ & $8.4 \cdot 10^{-220}$ & $-0.39$ & $2.5 \cdot 10^{-193}$ & $0.43$ & $1.3 \cdot 10^{-235}$ \\
        \cmidrule[0.5pt]{1-8}
    \end{tabular}
    \caption{Significance test results for the angular velocity experiments. The p-value approximately indicates the probability that the correlation coefficient is 0. As we can see, as angular velocity increases, the reconstruction quality decreases with low p-values.}
    \label{tab:angular_velocity_significance}
\end{table*}

%% file: figures/suppl/linear_angular_baselines_psnr_ssim.tex
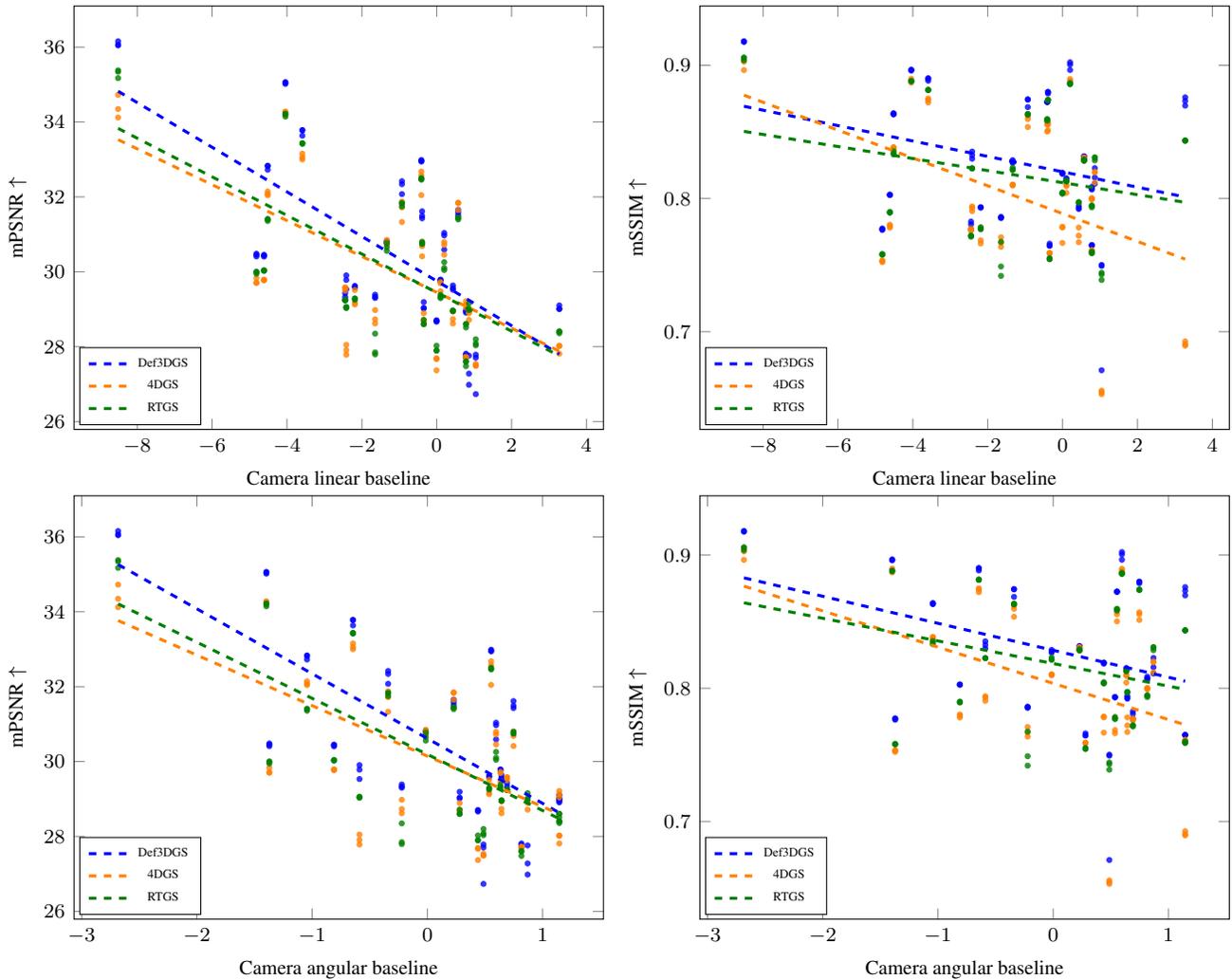
\begin{figure*}[t]
    \centering
    \begin{subfigure}[b]{0.5\textwidth}
        \centering
        \noindent
        \begin{tikzpicture}
        \begin{axis}[
            scale only axis,
            width=0.85\linewidth,
            height=0.85*0.8\linewidth,
            xlabel={Camera linear baseline},
            ylabel={mPSNR $\uparrow$},
            legend style={at={(0.01,0.01)},anchor=south west, font=\tiny},
            tick label style={font=\footnotesize},
            label style={font=\footnotesize}
        ]
        
        \begin{pgfonlayer}{background layer}
        \foreach \model/\color in \modelcolormapnoeg {
            \edef\temp{\noexpand\addplot[
                color=\color,
                x=baseline,
                y=lpips,
                mark=*,
                only marks,
                mark size=1.0pt,
                opacity=0.8,
                forget plot
            ]
            table[col sep=comma] {figures/suppl/psnr_baseline_csvs/linear_baseline_data_\model.csv};
            }
            
            \temp
        }
        \end{pgfonlayer}
        
        \begin{pgfonlayer}{foreground layer}
        \foreach \model/\color in \modelcolormapnoeg {
            \edef\temp{
            \noexpand\addplot[
                color=\color,
                x=baseline,
                y=lpips,
                dashed,
                line width=1.2pt
            ]
            table[col sep=comma] {figures/suppl/psnr_baseline_linear_csvs/linear_baseline_linear_data_\model.csv};
            \noexpand\addlegendentry{\model}
            }
            
            \temp
        }
        \end{pgfonlayer}
        
        \end{axis}
        \end{tikzpicture}
    \end{subfigure}%
    \begin{subfigure}[b]{0.5\textwidth}
        \centering
        \noindent
        \begin{tikzpicture}
        \begin{axis}[
            scale only axis,
            width=0.85\linewidth,
            height=0.85*0.8\linewidth,
            xlabel={Camera linear baseline},
            ylabel={mSSIM $\uparrow$},
            legend style={at={(0.01,0.01)},anchor=south west, font=\tiny},
            tick label style={font=\footnotesize},
            label style={font=\footnotesize}
        ]
        
        \begin{pgfonlayer}{background layer}
        \foreach \model/\color in \modelcolormapnoeg {
            \edef\temp{\noexpand\addplot[
                color=\color,
                x=baseline,
                y=lpips,
                mark=*,
                only marks,
                mark size=1.0pt,
                opacity=0.8,
                forget plot
            ]
            table[col sep=comma] {figures/suppl/ssim_baseline_csvs/linear_baseline_data_\model.csv};
            }
            
            \temp
        }
        \end{pgfonlayer}
        
        \begin{pgfonlayer}{foreground layer}
        \foreach \model/\color in \modelcolormapnoeg {
            \edef\temp{
            \noexpand\addplot[
                color=\color,
                x=baseline,
                y=lpips,
                dashed,
                line width=1.2pt
            ]
            table[col sep=comma] {figures/suppl/ssim_baseline_linear_csvs/linear_baseline_linear_data_\model.csv};
            \noexpand\addlegendentry{\model}
            }
            
            \temp
        }
        \end{pgfonlayer}
        
        \end{axis}
        \end{tikzpicture}
    \end{subfigure} \\
    \begin{subfigure}[b]{0.5\textwidth}
        \centering
        \noindent
        \begin{tikzpicture}
        \begin{axis}[
            scale only axis,
            width=0.85\linewidth,
            height=0.85*0.8\linewidth,
            xlabel={Camera angular baseline},
            ylabel={mPSNR $\uparrow$},
            legend style={at={(0.01,0.01)},anchor=south west, font=\tiny},
            tick label style={font=\footnotesize},
            label style={font=\footnotesize}
        ]
        
        \begin{pgfonlayer}{background layer}
        \foreach \model/\color in \modelcolormapnoeg {
            \edef\temp{\noexpand\addplot[
                color=\color,
                x=baseline,
                y=lpips,
                mark=*,
                only marks,
                mark size=1.0pt,
                opacity=0.8,
                forget plot
            ]
            table[col sep=comma] {figures/suppl/psnr_baseline_csvs/angular_baseline_data_\model.csv};
            }
            
            \temp
        }
        \end{pgfonlayer}
        
        \begin{pgfonlayer}{foreground layer}
        \foreach \model/\color in \modelcolormapnoeg {
            \edef\temp{
            \noexpand\addplot[
                color=\color,
                x=baseline,
                y=lpips,
                dashed,
                line width=1.2pt
            ]
            table[col sep=comma] {figures/suppl/psnr_baseline_linear_csvs/angular_baseline_linear_data_\model.csv};
            \noexpand\addlegendentry{\model}
            }
            
            \temp
        }
        \end{pgfonlayer}
        
        \end{axis}
        \end{tikzpicture}
    \end{subfigure}%
    \begin{subfigure}[b]{0.5\textwidth}
        \centering
        \noindent
        \begin{tikzpicture}
        \begin{axis}[
            scale only axis,
            width=0.85\linewidth,
            height=0.85*0.8\linewidth,
            xlabel={Camera angular baseline},
            ylabel={mSSIM $\uparrow$},
            legend style={at={(0.01,0.01)},anchor=south west, font=\tiny},
            tick label style={font=\footnotesize},
            label style={font=\footnotesize}
        ]
        
        \begin{pgfonlayer}{background layer}
        \foreach \model/\color in \modelcolormapnoeg {
            \edef\temp{\noexpand\addplot[
                color=\color,
                x=baseline,
                y=lpips,
                mark=*,
                only marks,
                mark size=1.0pt,
                opacity=0.8,
                forget plot
            ]
            table[col sep=comma] {figures/suppl/ssim_baseline_csvs/angular_baseline_data_\model.csv};
            }
            
            \temp
        }
        \end{pgfonlayer}
        
        \begin{pgfonlayer}{foreground layer}
        \foreach \model/\color in \modelcolormapnoeg {
            \edef\temp{
            \noexpand\addplot[
                color=\color,
                x=baseline,
                y=lpips,
                dashed,
                line width=1.2pt
            ]
            table[col sep=comma] {figures/suppl/ssim_baseline_linear_csvs/angular_baseline_linear_data_\model.csv};
            \noexpand\addlegendentry{\model}
            }
            
            \temp
        }
        \end{pgfonlayer}
        
        \end{axis}
        \end{tikzpicture}
    \end{subfigure}
    \caption{Camera linear and angular baseline results for mPSNR and mSSIM. As we can see, increasing any baseline leads to a visible drop in mPSNR. No such correlation can be observed for mSSIM.}
    \label{fig:baseline_psnr_ssim}
\end{figure*}

%% file: tables/suppl/baseline_signficance.tex
\begin{table*}[t]
    \centering
    \begin{tabular}{cccccccc}
        \cmidrule[0.5pt]{1-8}
        & & \multicolumn{2}{c}{\textbf{mPSNR} $\uparrow$} & \multicolumn{2}{c}{\textbf{mSSIM} $\uparrow$} & \multicolumn{2}{c}{\textbf{mLPIPS} $\downarrow$} \\
        \cmidrule(lr){3-4} \cmidrule(lr){5-6} \cmidrule(lr){7-8} 
        \textbf{Model} & \textbf{Coefficient} & Value & p-value & Value & p-value & Value & p-value \\
        \cmidrule{1-8}
        \multirow{2}{*}{Def3DGS} & Pearson & $-0.68$ & $3.1 \cdot 10^{-11}$ & $-0.29$ & $1.2 \cdot 10^{-2}$ & $0.24$ & $3.7 \cdot 10^{-2}$ \\
        & Spearman & $-0.64$ & $7.0 \cdot 10^{-10}$ & $-0.22$ & $5.6 \cdot 10^{-2}$ & $0.12$ & $2.9 \cdot 10^{-1}$ \\
        \cmidrule{1-8}
        \multirow{2}{*}{4DGS} & Pearson & $-0.61$ & $8.7 \cdot 10^{-9}$ & $-0.45$ & $6.3 \cdot 10^{-5}$ & $0.48$ & $1.6 \cdot 10^{-5}$ \\
        & Spearman & $-0.57$ & $8.5 \cdot 10^{-8}$ & $-0.32$ & $5.1 \cdot 10^{-3}$ & $0.34$ & $3.3 \cdot 10^{-3}$ \\
        \cmidrule{1-8}
        \multirow{2}{*}{RTGS} & Pearson & $-0.64$ & $4.3 \cdot 10^{-10}$ & $-0.24$ & $3.9 \cdot 10^{-2}$ & $0.13$ & $2.8 \cdot 10^{-1}$ \\
        & Spearman & $-0.57$ & $1.1 \cdot 10^{-7}$ & $-0.17$ & $1.5 \cdot 10^{-1}$ & $0.03$ & $8.2 \cdot 10^{-1}$\\
        \cmidrule[0.5pt]{1-8}
    \end{tabular}
    \caption{Significance test results for the linear baseline experiments. The p-value approximately indicates the probability that the correlation coefficient is 0. As we can observe, as the linear baseline increases, the reconstruction quality decreases but not all results are statistically significant  ($p \geq 0.05$).}
    \label{tab:linear_baseline_significance}
\end{table*}

\begin{table*}[t]
    \centering
    \begin{tabular}{cccccccc}
        \cmidrule[0.5pt]{1-8}
        & & \multicolumn{2}{c}{\textbf{mPSNR} $\uparrow$} & \multicolumn{2}{c}{\textbf{mSSIM} $\uparrow$} & \multicolumn{2}{c}{\textbf{mLPIPS} $\downarrow$} \\
        \cmidrule(lr){3-4} \cmidrule(lr){5-6} \cmidrule(lr){7-8} 
        \textbf{Model} & \textbf{Coefficient} & Value & p-value & Value & p-value & Value & p-value \\
        \cmidrule{1-8}
        \multirow{2}{*}{Def3DGS} & Pearson & $-0.72$ & $3.2 \cdot 10^{-13}$ & $-0.37$ & $1.0 \cdot 10^{-3}$ & $0.21$ & $6.8 \cdot 10^{-2}$ \\
        & Spearman & $-0.63$ & $1.6 \cdot 10^{-9}$ & $-0.25$ & $3.1 \cdot 10^{-2}$ & $0.04$ & $7.3 \cdot 10^{-1}$ \\
        \cmidrule{1-8}
        \multirow{2}{*}{4DGS} & Pearson & $-0.63$ & $1.9 \cdot 10^{-9}$ & $-0.43$ & $1.4 \cdot 10^{-4}$ & $0.35$ & $2.3 \cdot 10^{-3}$ \\
        & Spearman & $-0.52$ & $2.1 \cdot 10^{-6}$ & $-0.32$ & $5.6 \cdot 10^{-3}$ & $0.25$ & $3.0 \cdot 10^{-2}$\\
        \cmidrule{1-8}
        \multirow{2}{*}{RTGS} & Pearson & $-0.69$ & $9.2 \cdot 10^{-12}$ & $-0.33$ & $4.0 \cdot 10^{-3}$ & $0.16$ & $1.8 \cdot 10^{-1}$ \\
        & Spearman & $-0.54$ & $4.8 \cdot 10^{-7}$ & $-0.18$ & $1.2 \cdot 10^{-1}$ & $-0.03$ & $8.3 \cdot 10^{-1}$ \\
        \cmidrule[0.5pt]{1-8}
    \end{tabular}
    \caption{Significance test results for the angular baseline experiments. The p-value approximately indicates the probability that the correlation coefficient is 0. As we can observe, as the angular baseline increases, the reconstruction quality decreases but not all results are statistically significant ($p \geq 0.05$).}
    \label{tab:angular_baseline_significance}
\end{table*}

%% file: figures/angular_velocity.tex
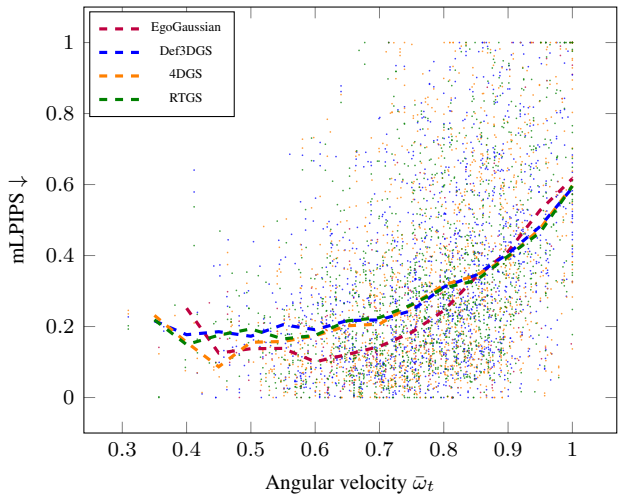
\begin{figure}[h]
    \centering
    \noindent
    \begin{tikzpicture}
    \begin{axis}[
        scale only axis,
        width=0.85\linewidth,
        height=0.85*0.8\linewidth,
        xlabel={Angular velocity $\bar{\omega}_t$},
        ylabel={mLPIPS $\downarrow$},
        xtick={0.3,0.4,0.5,0.6,0.7,0.8,0.9,1.0},
        ytick={0,0.2,0.4,0.6,0.8,1.0},
        legend style={at={(0.01,0.99)},anchor=north west, font=\tiny},
        tick label style={font=\footnotesize},
        label style={font=\footnotesize},
        title style={font=\footnotesize}
    ]
    
    \begin{pgfonlayer}{background layer}
    \foreach \model/\color in \modelcolormap {
        \edef\temp{\noexpand\addplot[
            color=\color,
            x=velocity,
            y=lpips,
            mark=*,
            only marks,
            mark size=0.1pt,
            opacity=0.8,
            forget plot
        ]
        table[col sep=comma] {figures/scatter_csvs/angular_scatter_data_\model.csv};
        }
        
        \temp
    }
    \end{pgfonlayer}
    
    \begin{pgfonlayer}{foreground layer}
    \foreach \model/\color in \modelcolormap {
        \edef\temp{
        \noexpand\addplot[
            color=\color,
            x=velocity,
            y=lpips,
            dashed,
            line width=1.2pt
        ]
        table[col sep=comma] {figures/trend_csvs/angular_trend_data_\model.csv};
        \noexpand\addlegendentry{\model}
        }
        
        \temp
    }
    \end{pgfonlayer}
    
    \end{axis}
    \end{tikzpicture}
    \caption{Camera angular velocity $\bar{\omega}_t$ plotted against mLPIPS. Additional trend lines are plotted. As we can observe, as angular velocity increases, mLPIPS increases, which corresponds to worse reconstruction quality.}
    \label{fig:angular_velocity}
\end{figure}

%% file: figures/angular_baseline.tex
\begin{figure}[t]
    \centering
    \noindent
    \begin{tikzpicture}
    \begin{axis}[
        title=Camera angular baseline against masked LPIPS,
        scale only axis,
        width=0.85\linewidth,
        height=0.85*0.8\linewidth,
        xlabel={Camera angular baseline},
        ylabel={mLPIPS $\downarrow$},
        legend style={at={(0.01,0.99)},anchor=north west, font=\tiny},
        tick label style={font=\footnotesize},
        label style={font=\footnotesize},
        title style={font=\footnotesize}
    ]
    
    \begin{pgfonlayer}{background layer}
    \foreach \model/\color in \modelcolormapnoeg {
        \edef\temp{\noexpand\addplot[
            color=\color,
            x=baseline,
            y=lpips,
            mark=*,
            only marks,
            mark size=1.0pt,
            opacity=0.8,
            forget plot
        ]
        table[col sep=comma] {figures/baseline_csvs/angular_baseline_data_\model.csv};
        }
        
        \temp
    }
    \end{pgfonlayer}
    
    \begin{pgfonlayer}{foreground layer}
    \foreach \model/\color in \modelcolormapnoeg {
        \edef\temp{
        \noexpand\addplot[
            color=\color,
            x=baseline,
            y=lpips,
            dashed,
            line width=1.2pt
        ]
        table[col sep=comma] {figures/baseline_linear_csvs/angular_baseline_linear_data_\model.csv};
        \noexpand\addlegendentry{\model}
        }
        
        \temp
    }
    \end{pgfonlayer}
    
    \end{axis}
    \end{tikzpicture}
    \caption{Camera angular baseline plotted against mLPIPS. Additional linear regression models fitted are shown. As we can observe, no clear correlation can be seen between camera angular baseline and mLPIPS.}
    \label{fig:angular_baseline}
\end{figure}
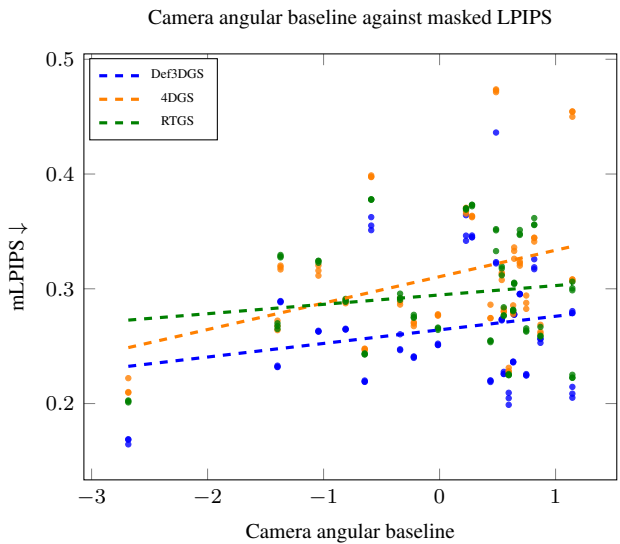